\documentclass[10pt,twocolumn,letterpaper]{article}

\usepackage[pagenumbers]{cvpr} 

\usepackage{graphicx}

\usepackage[table]{xcolor}
\definecolor{blue}{HTML}{0055cc}
\definecolor{red}{HTML}{cc1100}
\definecolor{orange}{HTML}{cc7700}
\definecolor{green}{HTML}{339955}
\definecolor{Highlight}{HTML}{39a58a}
\definecolor{LightGreen}{HTML}{39a58a}
\usepackage{cuted}
\usepackage{xcolor,soul}
\usepackage{capt-of}
\usepackage{etoc}
\usepackage{algorithm}
\usepackage{algorithmic}
\usepackage{mathtools}
\usepackage{tocloft}
\usepackage{afterpage}
\usepackage[inkscapeformat=png]{svg}
\usepackage[sectionbib]{chapterbib}
\usepackage[accsupp]{axessibility}
\usepackage{caption, subcaption, multirow, overpic, textpos}
\renewcommand{\paragraph}[1]{\vspace{0.75mm}\noindent\textbf{#1}}
\usepackage{array}          
\usepackage{natbib}

\newcolumntype{x}[1]{>{\centering\arraybackslash}p{#1pt}}
\newcolumntype{y}[1]{>{\raggedright\arraybackslash}p{#1pt}}
\newcolumntype{z}[1]{>{\raggedleft\arraybackslash}p{#1pt}}

\definecolor{cvprblue}{rgb}{0.21,0.49,0.74}
\usepackage{collectbox}
\usepackage[export]{adjustbox}
\usepackage{bbding}         
\usepackage{pifont} 

\definecolor{my_red}{HTML}{FE4444}
\definecolor{Highlight}{HTML}{39b54a}  
\definecolor{Gray}{gray}{0.95}
\definecolor{LightPurple}{HTML}{845071}
\definecolor{LightYellow}{HTML}{ECE0C2}
\definecolor{MiddleBrown}{HTML}{887F7F}
\definecolor{MiddlePurPle}{HTML}{81506E}
\definecolor{LightGreen}{HTML}{dff0ea}

\usepackage[pagebackref,breaklinks,colorlinks,linkcolor=purple,citecolor=purple,urlcolor=purple]{hyperref} 

\def\paperID{} 
\def\confName{}
\def\confYear{}

\title{Prompt Highlighter: Interactive Control for Multi-Modal LLMs}

\author{Yuechen Zhang$^{1}$\hspace{1.0cm}Shengju Qian$^{1}$\hspace{1.0cm}Bohao Peng$^{1}$\hspace{1.0cm}Shu Liu$^{2}$\hspace{1.0cm}Jiaya Jia$^{1,2}$\\
$^{1}$The Chinese University of Hong Kong~~~
$^{2}$SmartMore\\
\small
{\small\textbf{\url{https://github.com/dvlab-research/Prompt-Highlighter}}}
\vspace{-10pt}
}

\begin{document}
\maketitle
\begin{strip}\centering
    \vspace{-35pt}
    \captionsetup{type=figure}
    \includegraphics[width=1\textwidth]{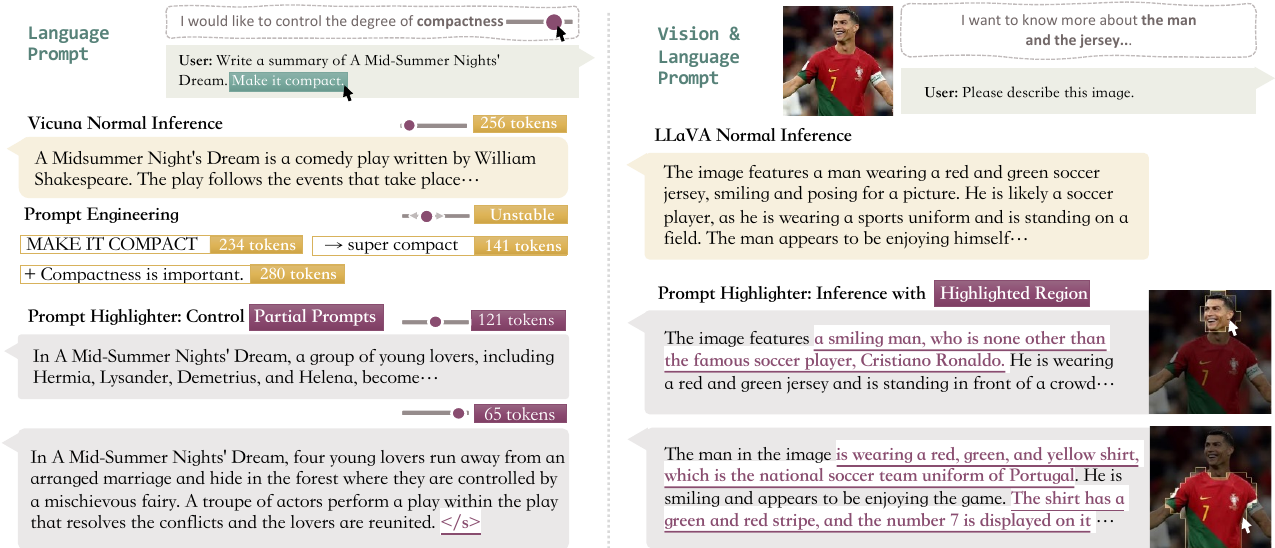}
    \vspace{-15pt}
    \captionof{figure}{Prompt Highlighter facilitates token-level user interactions for customized generation, compatible with both LLMs and VLMs. Compared with \sethlcolor{LightYellow}\textcolor{black}{\hl{ vanilla inference }} and prompt engineering, the \sethlcolor{Gray}\textcolor{purple}{\hl{ context-highlighted inference }} provided by our method offers controllable generations and produces customized results. Outputs correlated with the highlighted parts are \underline{underlined}.
    }\label{fig:teaser_image}
\end{strip}

\vspace{-5pt}

\begin{abstract}
\vspace{-3mm}

This study targets a critical aspect of multi-modal LLMs' ~(LLMs\&VLMs) inference: explicit controllable text generation. Multi-modal LLMs empower multi-modality understanding with the capability of semantic generation yet bring less explainability and heavier reliance on prompt contents due to their autoregressive generative nature. While manipulating prompt formats could improve outputs, designing specific and precise prompts per task can be challenging and ineffective. To tackle this issue, we introduce a novel inference method, Prompt Highlighter, which enables users to highlight specific prompt spans to interactively control the focus during generation. Motivated by the classifier-free diffusion guidance, we form regular and unconditional context pairs based on highlighted tokens, demonstrating that the autoregressive generation in models can be guided in a classifier-free way. Notably, we find that, during inference, guiding the models with highlighted tokens through the attention weights leads to more desired outputs. Our approach is compatible with current LLMs and VLMs, achieving impressive customized generation results without training. Experiments confirm its effectiveness in focusing on input contexts and generating reliable content. Without tuning on LLaVA-v1.5, our method secured 70.7 in the MMBench test and 1552.5 in MME-perception. 

\vspace{-3mm}
\end{abstract}
\vspace{-3mm}

\section{Introduction}\label{sec:intro}
\vspace{-2mm}
Large Language Models (LLMs) have driven significant progress in a multitude of natural language processing tasks~\cite{radford2018improving,radford2019language,brown2020language,liu2019roberta,devlin2018bert,lewis2019bart,touvron2023llama,chiang2023vicuna,chung2022scaling}. 
Further advancements have been achieved by extending these models to handle vision-language tasks~\cite{liu2023visual,zhu2023minigpt,li2022blip,li2023blip,ye2023mplug,dai2023instructblip} through visual-language alignment and instruction tuning. These efforts have led to the development of Vision Language Models (VLMs), which can generate text based on multi-modal inputs.
Due to its autoregressive nature, the typical generation process in LLMs and VLMs~(multi-modal LLMs) is primarily conditioned on input contexts. Prompt engineering~\cite{yao2023tree,wei2022chain,Significant-Gravitas,wang2022self} has emerged as a common interaction mechanism between humans and language models, where diverse formats and content of prompts are employed to steer the generation towards desired outcomes. However, prompt engineering often relies on empirical intuition and requires careful design of the context, making it less accessible for non-experts. As illustrated in the left part of~\cref{fig:teaser_image}, 
even the meticulously crafted prompts, which convey the concept of `compactness' clearly, can lead to unpredictable outputs that fail to meet the requirements.

Instead of manipulating prompt-level contexts (\ie, prompt engineering) to control LMs' generation process, we propose a novel inference approach, Prompt Highlighter, that enables token-level user interactions for personalized generations.
Our method allows users to interact with multi-modal LLMs in a manner analogous to applying a highlighter tool on the input context in the text editor, enabling users to emphasize desired parts by highlighting them. 

This highlighting mechanism is achieved by constructing a regular and unconditional input context pair with different textual embeddings in the highlighted tokens. Subsequently, we can adjust the model's focus on the highlighted components by employing the classifier-free guidance~\cite{ho2022classifier,sanchez2023stay,kornblith2023guiding} on predicted token probabilities.
Moreover, by probing cross-token attention maps, we discover a robust correlation between attention scores and the semantic significance of tokens. This suggests that, in the autoregressive generation process of language models, the semantic relationship between tokens can be represented to a certain extent by their attention scores. Building on this insight, we introduce an attention activation strategy that adjusts the attention weights associated with a highlighted part.
Specifically, Prompt Highlighter employs an adjusted attention mask to reweight corresponding attention scores, enabling a more focused generation on highlighted parts. 

As illustrated in~\cref{fig:teaser_image}, compared to vanilla inference, our highlighted inference can guide the generation process to produce controllable results that align more closely with user needs. Prompt Highlighter is compatible with mainstream transformer-based multi-modal LLMs. This compatibility encompasses VLMs that use precise patch-wise visual token mapping, such as LLaVA~\cite{liu2023visual,liu2023improved,fuyu-8b}, as well as methods that employ implicit query-based visual token mapping, like those based on Q-Former~\cite{li2023blip,dai2023instructblip,ye2023mplug,zhu2023minigpt}. This novel interaction paradigm with highlighted sections during the generation process goes beyond what prompt engineering can offer.

We further demonstrate the effectiveness of Prompt Highlighter by evaluating it using comprehensive multi-modal benchmarks. We verify that directly highlighting the full image context in VLMs can significantly improve the quality of generated image captions~\cite{lin2014microsoft} and question-answering results. Specifically, our method can effectively mitigate the model's propensity to hallucinate by guiding its focus toward reliable contexts, thereby enhancing overall performance. Notably, without additional training, our method improves the performance of the baseline LLaVA-v1.5, securing 2$^{\text{nd}}$ place in both MMBench~\cite{liu2023mmbench} and MME-perception~\cite{fu2023mme} leaderboards.

Our contributions can be summarized as follows:
(1) We pioneer the exploration of fine-grained human-model interactions in multi-modal LLMs, proposing a plug-and-play pipeline that enables token-level user interactions for controllable generation.
(2) We conduct extensive experiments on comprehensive benchmarks, demonstrating that our method significantly enhances the overall performance.

\vspace{-3mm}
\section{Related Works}\label{sec:related_work}
\vspace{-2mm}
\subsection{Multi-Modal LLMs}
\label{sec:related_vlvm}
\vspace{-2mm}
Recent Large Language Models (LLMs) ~\cite{touvron2023llama,touvron2023llama2,chiang2023vicuna,chowdhery2022palm,radford2021learning,radford2018improving,chung2022scaling} play a significant role in natural language processing tasks, particularly in language generation and question answering.
Building upon these pre-trained language models, Vision-Language Models~(VLMs) ~\cite{liu2023visual,ye2023mplug,peng2023kosmos,li2023blip,zhu2023minigpt,dai2023instructblip} further introduce the alignment between vision and language modalities by leveraging extensive training on image-caption pairs or image-question conversations.
There are two prevalent methods for aligning vision and verbiage modalities. The first method, exemplified by LLaVA~\cite{liu2023visual}, directly maps image patches to tokens using a projector, establishing a one-to-one correspondence. The second method, represented by models like BLIP2~\cite{li2023blip,zhang2023internlmxcomposer}, employs a Query Transformer (Q-Former) after getting image features to establish a non-uniform patch-token mapping. These methods use learnable queries to get compressed image features, yielding visual tokens rich with semantic information.

\vspace{-2mm}
\subsection{Interactions with Multi-Modal LLMs}
\vspace{-2mm}
\paragraph{Prompt engineering and interactions.} Based on the autoregressive property of LLMs, users aim to control the generation results by modifying the input contexts. This largely determines the test-time interactions with LLMs, primarily executed through prompt engineering. Representative methods such as CoT~\cite{wei2022chain} introduce demonstrations in the context to enhance reasoning ability. Other multi-branch designs like ToT and GoT~\cite{yao2023tree,besta2023graph,Significant-Gravitas,xie2023decomposition,wang2022self} have been proposed for rich and reliable context generation and self-checking.
Aside from prompt engineering, human-model interactions have not been extensively explored in VLMs. Methods like Kosmos-2~\cite{peng2023kosmos}, LLaVAInteractive~\cite{chen2023llavainteractive}, LISA~\cite{lai2023lisa}, and AlphaCLIP~\cite{sun2023alphaclip} enable grounding perception tasks such as detection, segmentation, caption, and image editing through interaction with LLMs. These task-oriented interactions require additional data collection and task-specific tuning.
In contrast, Prompt Highlighter is plug-and-play for general text generation in pre-trained models.

\begin{figure*}[ht!]
    \centering
    \includegraphics[width=0.99\linewidth]{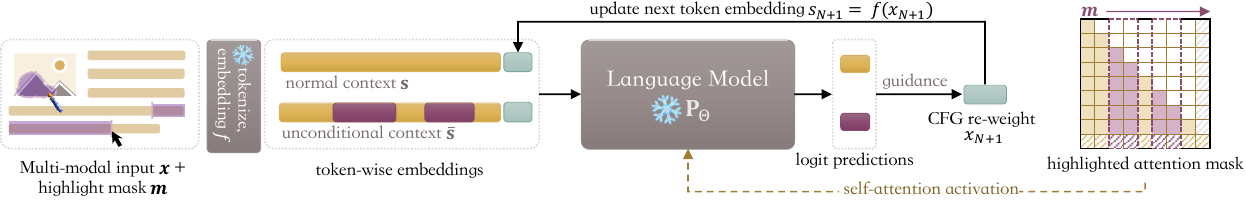}
    \vspace{-2mm}
    \caption{An abstract pipeline of Prompt Highlighter. Users can control the focus of generation by marking out specific image regions or text spans. Then a token-level mask $\mathbf{m}$ is created to guide the language model's inference.}\label{fig:pipeline}
    \vspace{-5mm}
\end{figure*}

\paragraph{Classifier-free guidance and controllable generation.}
Classifier-Free Guidance (CFG) ~\cite{ho2022classifier} enables a control on Diffusion Models' generation process without a conventional classifier. Specifically, CFG's step-wise sampling allows users to employ a negative prompt within the unconditional branch, effectively guiding the generation away from harmful distributions. This approach has been extended to language models by LLM-CFG ~\cite{sanchez2023stay}, allowing a controllable text generation and improved performance. However, LLM-CFG still requires a pair-wise prompt design and does not support partial token-level reweighting within the context, which is vital for controlling VLM's generation. Besides, methods in Diffusion Models ~\cite{ge2023expressive,chefer2023attend} achieve fine-grained control over image generation using text prompts by emphasizing areas within cross-attention maps. Fine-grained control over autoregressive generation in LLMs and VLMs is still challenging. Later concurrent works CRG and MARINE~\cite{Wan2024CRG, zhao2024mitigating}, adopt CFG in VLMs for grounding and mitigating hallucination, but employ a different design for positive-negative pairs compared to our approach.

\vspace{-2mm}
\section{Prompt Highlighter}\label{sec:method}
\vspace{-2mm}
An overview of Prompt Highlighter is presented in \cref{fig:pipeline}. Given a pre-trained generative model $\mathbf{P}_{\Theta}$, we first extract the input tokens from the text and the input image, forming the prompt context $\mathbf{x}$. Subsequently, by marking out specific image regions
or text spans, user creates a token-level binary mask $\mathbf{m}$ to highlight specific tokens.
Prompt Highlighter then generates the output sequence $\mathbf{y}$ using the two-branch condition based on $\mathbf{m}$ autoregressively. The following section will delve into the specifics of our method.

\vspace{-1mm}
\subsection{Token-Level Highlight Guidance}\label{sec:highlight_method}
\vspace{-1mm}
In conditioned Diffusion Models ~\cite{dhariwal2021diffusion}, given a noisy image $x$ and a class condition $c$, the model predicts probability likelihood $\hat{\mathbf{P}}$, for the conditioned step-wise sample, $\hat{\mathbf{P}}_{\Theta}(x|c)\varpropto \mathbf{P}_{\Theta}(x)\cdot \mathbf{P}_{\Phi}(c|x)^{\gamma}$. Here, $\mathbf{P}_{\Phi}$ is a classifier, and $\gamma$ is the guidance strength controlling the weight of likelihood on $c$. Ho et al. ~\cite{ho2022classifier} observed that guidance can be offered without a classifier. Applying the Bayes rule, $\mathbf{P}_{\Theta}(c|x) \varpropto \mathbf{P}_{\Theta}(x|c)/\mathbf{P}_{\Theta}(x)$, the sampling process of the Classifier-Free Guidance~(CFG) can be expressed as 
\begin{equation}\label{eq:cfg_diffusion}
    \mathbf{\hat{P}}_{\Theta}(x|c) \varpropto \mathbf{P}_{\Theta}(x|c)^\gamma /\mathbf{P}_{\Theta}(x)^{\gamma-1}\text{,}
\end{equation}
\vspace{-4mm}
\begin{align}\label{eq:cfg_diffusion_prob}
\log\mathbf{\hat{P}}_{\Theta}(\epsilon_t|x_{t+1},c)
& = \gamma\log\mathbf{P}_{\Theta}(\epsilon_t|x_{t+1},c) \\
& - (\gamma-1)\log\mathbf{P}_{\Theta}(\epsilon_t|x_{t+1})\text{,} \nonumber
\vspace{-1mm}
\end{align}
in which $\epsilon_t$ is the noise prediction conditioned on the previous output $x_{t+1}$ and the text condition $c$. LLM-CFG~\cite{sanchez2023stay} extended this property to autoregressive language models. Given a sequence of $N$ tokens $\mathbf{x} = \{x_1,\ldots,x_N\}$, the likelihood of predicting the entire sequence can be expressed as $\mathbf{P}_{\Theta}(x)=\prod^N_i \mathbf{P}_{\Theta}(x_i | x_{j<i})$. The model samples each subsequent token from the conditional probability distribution. Based on~\cref{eq:cfg_diffusion}, the CFG sampling on the language model can be denoted as
\vspace{-1mm}
\begin{equation}\label{eq:cfg_lang}
    \mathbf{\hat{P}}_{\Theta}(\mathbf{x}|c) \varpropto \frac{\mathbf{P}_{\Theta}(\mathbf{x}|c)^\gamma}{\mathbf{P}_{\Theta}(\mathbf{x})^{\gamma-1}} \varpropto \prod^N_{i=1} \frac{\mathbf{P}_{\Theta}(x_i | x_{j<i}, c)^\gamma}{ \mathbf{P}_{\Theta}(x_i | x_{j<i})^{\gamma-1}} \text{.}
\end{equation}
Similar to the transaction from~\cref{eq:cfg_diffusion} to~\cref{eq:cfg_diffusion_prob}, the likelihood in LLM is represented as the next-token classification probability. Thus next token's logit prediction $\mathbb{P}{x_i} = \log\mathbf{\hat{P}}_{\Theta}(x_i|x_{j<i}, c)$ is
\begin{equation}\label{eq:cfg_lang_prob}
    \mathbb{P}{x_i} = \gamma\log\mathbf{P}_{\Theta}(x_i|x_{j<i}, c) - (\gamma-1)\log\mathbf{P}_{\Theta}(x_i|x_{j<i})\text{.}
    \vspace{-1mm}
\end{equation}

The formulation in~\cref{eq:cfg_lang,eq:cfg_lang_prob} offers a paradigm for controllable generation in LLMs~\cite{sanchez2023stay}, with the guidance strength $\gamma$ controls the degree of generation focus. Notably, the effectiveness of this guidance depends on the careful design of the conditional prompt $c$, which should be naturally formed as a complete phrase or sentence to retain its semantic meaning.
Prompt Highlighter extends CFG control in language models in a more generalized manner. The user's selection on the context $\mathbf{x}$ is converted into a token-level binary highlight mask $\mathbf{m}=\{m_1,\ldots,m_N\}$. We define $m_i=1$ if the $i$-th token $x_i$ is highlighted, and $m_i=0$ otherwise. This mask constructs a two-branch condition: the normal and the unconditional contexts.
The normal context operates in the same manner as in vanilla inference. Meanwhile, the unconditional context $\bar{\mathbf{s}}$ is derived from the normal conditional context $\mathbf{s}=\{s_1,\ldots,s_N\}$ within the textual embedding space through a token-wise scaling,
\begin{figure*}[t!]
    \centering
     \includegraphics[width=1\linewidth]{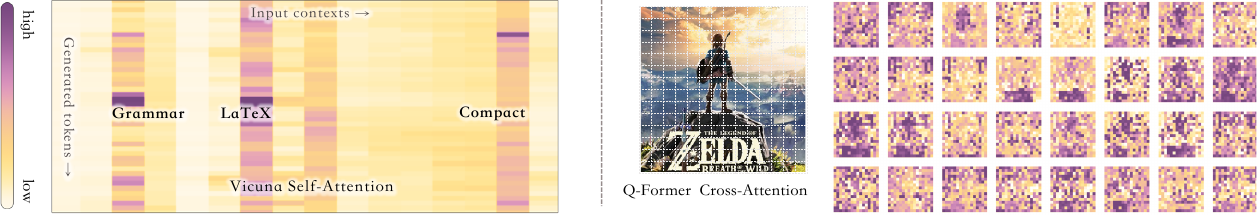}
     \vspace{-3mm}
     \caption{Visualizing attention maps. \textbf{\textit{Left}}: A segment of the attention map between the generated tokens and the input requirement prompt: ``\ldots fix the grammar and keep LaTeX format, make it compact\ldots''. Some representative tokens are marked for reference. \textbf{\textit{Right}}: Query-based token mapping. This shows the attention score on 32 queries in the first cross-attention layer of the Q-Former.}\label{fig:attn_vis}
     \vspace{-4mm}
\end{figure*}
\begin{equation}\label{eq:cfg_embed}
\bar{s}_i = (\alpha - 1)m_i \cdot f(x_i) + f(x_i)\text{,}
\end{equation}
where $\alpha$ is the scaling factor and $f(\cdot)$ is the token-to-embedding function, \ie, ${s}_i = f(x_i)$. We empirically set a small rescale $\alpha$ (\eg, 0.01) that can ensure a normal inference while ignoring the highlighted part. Then, based on the two-branch condition $(\mathbf{s}, \bar{\mathbf{s}})$, we can define the $i$-th token sampling process of the token-level highlight guidance as
\vspace{-1mm}
\begin{align}\label{eq:cfg_hl_prob}
    \log\mathbf{\hat{P}}_{\Theta}(x_i|s_{j<i}) = 
    & \gamma\log\mathbf{P}_{\Theta}(x_i|s_{j<i}) \nonumber\\
    & - (\gamma-1)\log\mathbf{P}_{\Theta}(x_i|\bar{s}_{j<i})\text{.}
    \vspace{-2mm}
\end{align}
Compared with~\cref{eq:cfg_lang_prob}, the additional conditional context $c$ is naturally incorporated as the difference between $\mathbf{s}$ and $\bar{\mathbf{s}}$. This arrangement provides users with the flexibility to control the in-line requirements. After the $i$-th token is predicted, the highlight mask $\mathbf{m}$ and contexts $\mathbf{s}, \bar{\mathbf{s}}$ are updated by appending $m_i=0$ and ${s}_i=\bar{s}_i = f(x_i)$, respectively. The generation process is terminated when the end token \texttt{</s>} is predicted.

\vspace{-1mm}
\subsection{Attention Activation}\label{sec:attention_activation}
\vspace{-1mm}
The token-level highlight guidance anchors the generative process with a token-wise logit reweighting. However, its effectiveness may diminish when facing long and complex input contexts with few highlighted tokens, making it difficult to distinguish $\mathbf{s}$ and $\mathbf{\bar{s}}$. To further investigate token-wise correlations and their impact on generation results, we exclude sink tokens that dominate the attention score~\cite{xiao2023streamingllm} and visualize cross-token self-attention score maps during inference.
For instance, in the left of~\cref{fig:attn_vis}, pivotal tokens form a band-like pattern on the attention map, drawing attention from nearly all following tokens. This pattern endures with changes in the model's layer number or attention heads, suggesting the attention mechanism's consistency and robustness in well pre-trained LLMs~\cite{chiang2023vicuna,touvron2023llama}. Meanwhile, it implies that attention scores within the model can represent the semantic correlation between tokens.

When addressing diverse requirements, LLMs need to balance attention among multiple tokens. For instance, as seen in~\cref{fig:attn_vis} (left), as one of the requirements in the prompt, `compactness' might not get enough attention during the generation process, resulting in an output that is less compact than expected.
Given the direct correlation observed between attention and tokens, we propose an attention activation strategy to activate the attention scores on highlighted tokens within the attention mechanism. This strategy can effectively steer the model's focus towards or away from specific tokens, allowing for more nuanced and precise control over the output.
We reformulate the model inference function in~\cref{eq:cfg_hl_prob} to a mask-conditioned one $\tilde{{\mathbf{P}}}_{\Theta}(x_i|s_{j<i}, \mathbf{m})$. In each of its self-attention layers, let $\mathbf{k}_i$ represent the $i$-th column vector of the query-key multiplied attention score matrix in one attention head. The activated attention score $h_i$ \textbf{\textit{in the normal context branch}} is defined as
\vspace{-1mm}
\begin{equation}\label{eq:attntion_add}
    h_i = \log(\beta)\cdot m_i + k_i\text{.}
\end{equation}
Then, the attention probability ${p}_i$ is calculated as
\begin{equation}\label{eq:attntion_explain}
    {p}_i = \frac{\exp(h_i)}{\sum_{j=1}^N \exp(h_j)}=\frac{\beta^{m_i}\cdot\exp(k_i)}{\sum_{j=1}^N \beta^{m_j}\cdot\exp(k_j)}\text{.}
\end{equation}
This mechanism defines the activation scaling factor as $\beta$. \textbf{\textit{For the unconditional branch}}, the attention score is deactivated by using a scaled negative mask in the inference $\tilde{{\mathbf{P}}}_{\Theta}(x_i |\bar{s}_{j<i}, -\delta\mathbf{m})$.~\cref{eq:attntion_explain} presents a SoftMax probability ${p}_i = \operatorname{softmax}(h_i)$ on the activated scores, with a consistent $\beta^{m_i}$-times probability augmentation on highlighted tokens. The attention activation operates under the assumption that users cannot highlight dominant `sink' tokens as explored in~\cite{xiao2023streamingllm}. Consequently, the attention activation will not catastrophically impair the model's fundamental generative capabilities during the inference.

\subsection{Highlighting Visual Tokens}\label{sec:vision_highlight}

Methods for highlighting visual tokens can be classified into two categories based on the type of token mapping involved, as discussed in~\cref{sec:related_vlvm}.
In direct token mapping, such as in LLaVA~\cite{liu2023visual}, the highlighting of visual tokens is straightforward. Image patch-level feature forms the sequential visual contexts $\mathbf{s^{\text{im}}}$. This enables a natural patch-level scaling on embeddings in~\cref{sec:highlight_method} and attention activation introduced in~\cref{sec:attention_activation}.

\begin{figure}[t!]
    \centering
     \includegraphics[width=0.95\linewidth]{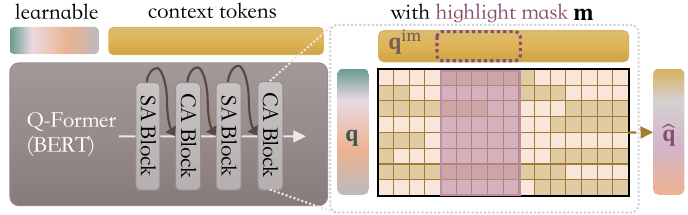}
     \vspace{-2mm}
     \caption{Highlighting visual tokens with Q-Former-based methods. In comparison with vanilla inference, we augment the learnable queries $\mathbf{q}$ by activating corresponding attention weights in the Cross-Attention~(CA) blocks.}\label{fig:qformer}
     \vspace{-2mm}
\end{figure}

In contrast, the scenario becomes more complex with query-based token mapping. For example, in works like BLIP-2~\cite{li2023blip,zhu2023minigpt,ye2023mplug}, the image feature is transformed into a unified set of few learnable queries $\mathbf{q}$ via Q-Former, which are then input into LLMs as textual embeddings. This process obscures the direct correlation between image patches and input tokens, as demonstrated in the example on the right side of~\cref{fig:attn_vis}.
To address this challenge, we leverage the fact that the Q-Former itself is a Transformer model to perform the token highlighting directly inside it. First, we adopt the embedding rescale in~\cref{eq:cfg_embed} on the patch-wise image feature and use the output queries from the Q-Former to form the image context pair $(\mathbf{s}^{\text{im}}, \mathbf{\bar{s}}^{\text{im}})$. Then, by activating attention scores within the corresponding patch-wise user-selection mask $\mathbf{m}$ in Q-Former's cross-attention layers, we can effectively steer the model to adjust its focus.
This process is depicted in~\cref{fig:qformer}. In the Q-Former model, cross attention is calculated across the learnable query $\mathbf{q}=\{{q_1,\ldots,q_M}\}$ and the image feature $\mathbf{q}^{\text{im}}=\{{q^{\text{im}}_1,\ldots,q^{\text{im}}_N}\}$. We then activate the attention score corresponding to the mask $\mathbf{m}$ within the cross-attention map. It can be expressed using a formulation similar to~\cref{eq:attntion_add},

\begin{figure}[t!]
    \centering
     \includegraphics[width=0.95\linewidth]{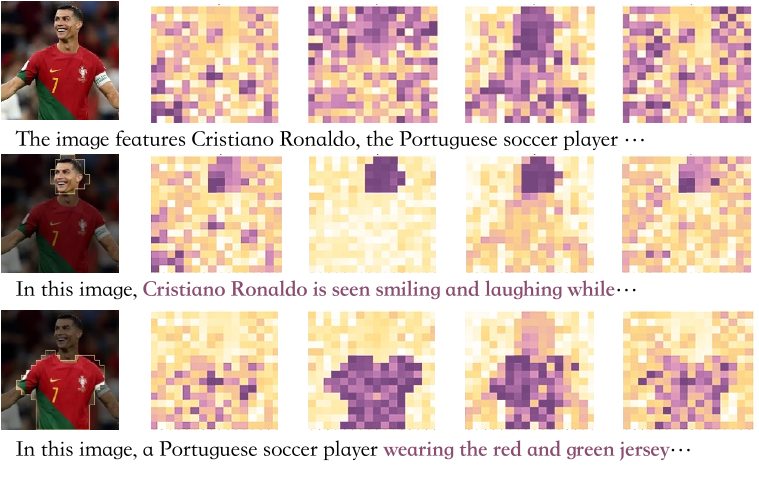}
     \vspace{-3mm}
     \caption{Attention scores in the first four queries of the Q-Former. Each row shows a different user selection and text output.}\label{fig:qformer_vis}
     \vspace{-4mm}
\end{figure}

\vspace{-2mm}
\begin{equation}\label{eq:qformer}
    \mathbf{\hat{q}} = \operatorname{softmax}\left(({\log(\beta)\cdot \mathbf{w} + QK^\top})/{\sqrt{d_k}}\right)V\text{,}
    \vspace{-1mm}
\end{equation}
where $Q=f_q(\mathbf{q})$, $K=f_k(\mathbf{q}^{\text{im}})$, and $V=f_v(\mathbf{q}^{\text{im}})$ represent the transformed QKV vectors in the cross attention with projections $f_{(\cdot)}$, and $d_k$ denotes the dimension of $K$. As a variation of the activation process in~\cref{eq:attntion_add,eq:attntion_explain}$, \mathbf{w}$ is a resized version of $\mathbf{m}$, expanding dimension from $N$ to $(H, M, N)$, where $H$ denotes the number of attention heads.
We illustrate the variations in attention scores with changing input masks in~\cref{fig:qformer_vis}. Attention score maps show attention concentrations in masked areas. This guides the model's focus towards specific visual elements, leading to various generation outputs.

\begin{figure*}[ht!]
    \centering
    \includegraphics[width=0.95\linewidth]{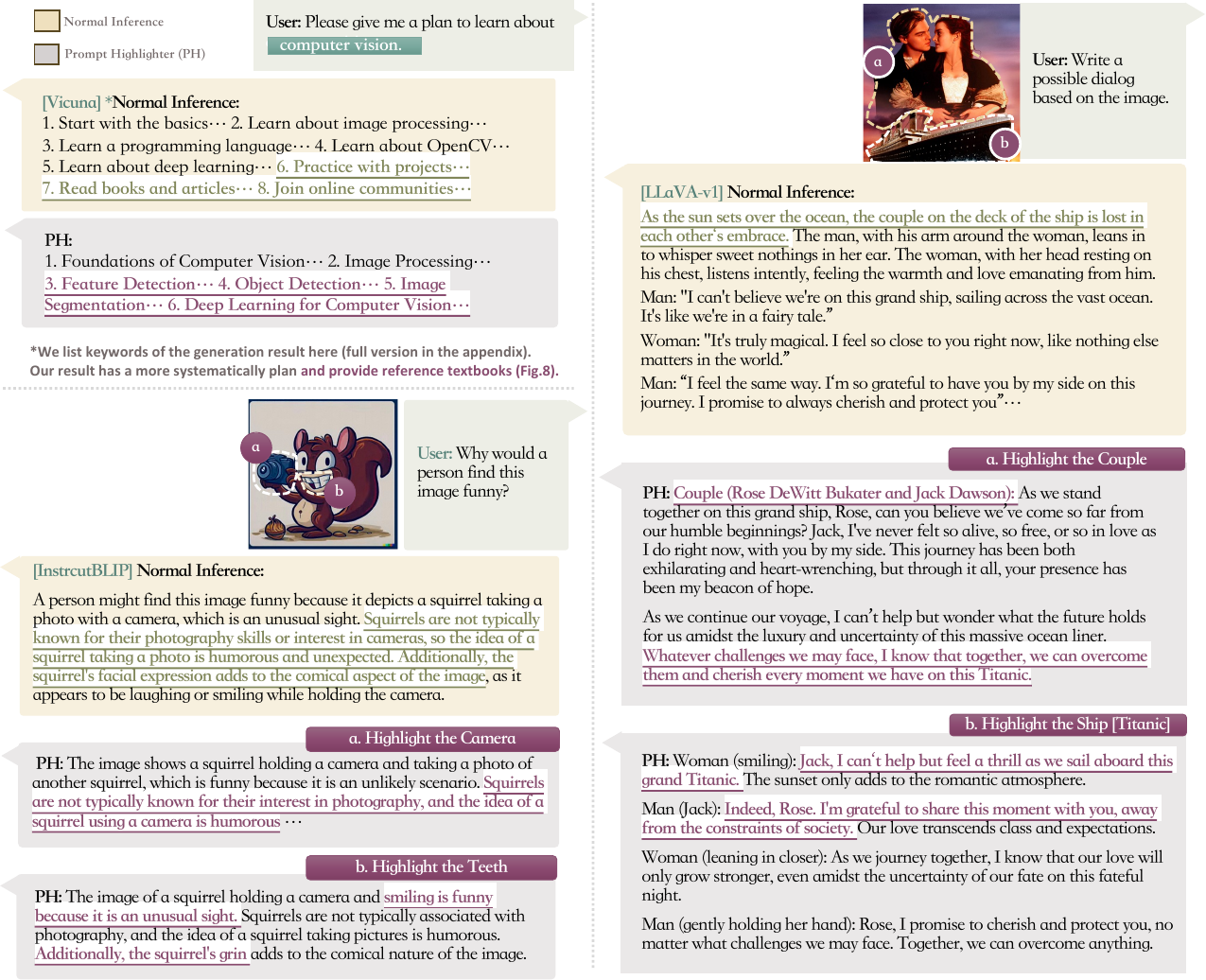}
    \caption{Partial context highlighter. When given \sethlcolor{LightGreen}\hl{ user input }, \sethlcolor{LightYellow}\hl{ vanilla inference } might lead to unfocused generations. Our \sethlcolor{Gray}\textcolor{purple}{\hl{ context-highlighted inference }} can faithfully capture the content of the highlighted part. The highlighted sections in each case are marked with circled indices in the image. Outputs correlated with the highlighted parts are \underline{underlined}.}\label{fig:showcase}
    \vspace{-4mm}
\end{figure*}

\begin{figure*}[ht!]
    \centering
    \includegraphics[width=1\linewidth]{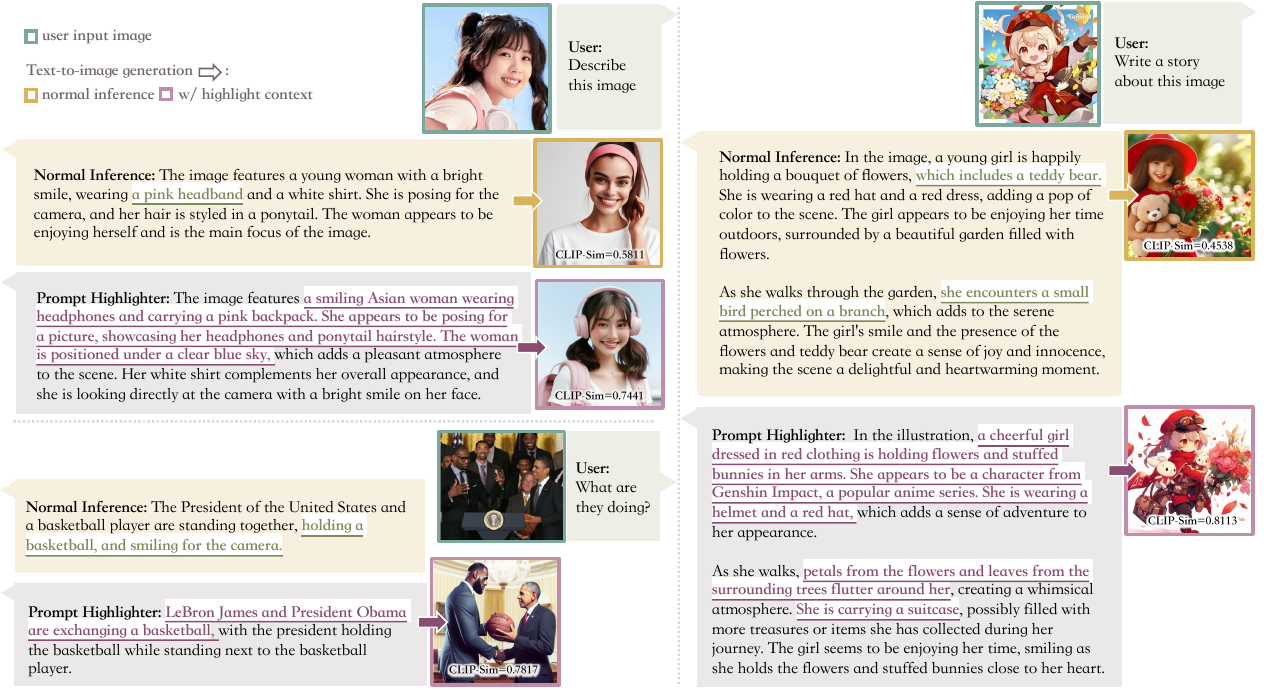}
    \vspace{-5mm}
    \caption{Results when highlighting all input contexts. Given \sethlcolor{LightGreen}\hl{ user input }, \sethlcolor{LightYellow}\hl{ vanilla inference } might lead to hallucinations. In contrast, \sethlcolor{Gray}\textcolor{purple}{\hl{ context-highlighted inference }} can accurately capture the content of the image. We further feed descriptions into DALLE-3~\cite{betker2023improving} (shown on the right) to provide a visually apparent difference. The CLIP-Similarity~\cite{radford2021learning} with the input image is reported for each generated image.} \label{fig:applications}
    \vspace{-2mm}
\end{figure*}

\vspace{-1mm}
\section{Experiments}\label{sec:experiments}
\vspace{-1mm}
\subsection{Implementation Details}\label{sec:impl_details}
\vspace{-1mm}
\paragraph{Pre-trained models.} Prompt Highlighter can be applied to a variety of general frameworks. For LLMs, we employ the LLaMA model architecture~\cite{touvron2023llama} and utilize Vicuna-13B v1.1~\cite{chiang2023vicuna} as the test model. For VLMs, experiments are done on one direct token mapping model LLaVA-13B~\cite{liu2023visual,liu2023improved} and one query-based mapping model InstructBLIP-Vicuna-13B~\cite{dai2023instructblip}. We adopt the LLaVA-v1.5 13B model~\cite{liu2023improved} in quantitative evaluations. All experiments are conducted on a single NVIDIA-A800 GPU.

\paragraph{Hyper-parameters.} The following parameter configurations are constant throughout almost all examples. In the highlight guidance, we set the guidance strength $\gamma=1.3$ in~\cref{eq:cfg_hl_prob}, and the scaling parameter $\alpha$ in~\cref{eq:cfg_embed} is set to $0.01$. In attention activation, we accommodate the diverse feature domains across different models by setting $\beta=2.0$ in~\cref{eq:attntion_add} and $\beta=20.0$ in~\cref{eq:qformer}. The scale factor $\delta$ in~\cref{eq:attntion_add} is set to satisfy $\delta\mathbf{m} = (\log(\beta) + 2)\mathbf{m}$. 

\paragraph{Inference.} During user interactions, if the text range selected by the user does not align perfectly with the tokens from the tokenizer, we adjust the start or end selection position to ensure that the selected range is fully encompassed. For images, the input mask is downsampled to a patch-wise binary mask based on the input image. The size depends on the visual encoder's patch size (e.g., CLIP).
In the autoregressive generation process, we employ a greedy search and cache prior KV values across layers.

\vspace{-1mm}
\subsection{Applications and Comparisons}\label{sec:applications}
\vspace{-1mm}
\paragraph{Partial context highlighting} is a fundamental application of Prompt Highlighter, utilized in scenarios where the emphasis is required on specific spans of the context. As shown in~\cref{fig:teaser_image,fig:showcase}, our method enables users to guide the model's attention toward relevant parts of the input by highlighting them. It enhances the focus and relevance of the generated output across a range of diverse frameworks~\cite{chiang2023vicuna,dai2023instructblip,liu2023visual}, making results align more precisely with user-selected tokens. It's worth noting that the examples in~\cref{fig:showcase} can not be achieved by LLM-CFG~\cite{sanchez2023stay}, as it requires a prompt design that has complete sentences or images.
This approach proves particularly useful in tasks such as content summarization and interactive conversations.

\paragraph{Generation control.} As demonstrated in~\cref{fig:teaser_image}, we can further control the degree of correlation to the highlighted part in text generation. Users can manipulate the model's output dynamically by adjusting the highlight guidance strength $\gamma$ in~\cref{eq:cfg_hl_prob}. This capability can prove advantageous in a variety of tasks, ranging from generating descriptive captions for images to creating customized responses in conversational agents.

\paragraph{Reliable description.} When carrying out long text generation tasks, every predicted token interacts with all previous content, including the newly generated contexts, to gather information. This may lead to a gradual divergence and loss of attention between the generated tokens and the input context, causing the model to hallucinate. We address this issue by highlighting all user input condition tokens, thereby guiding the generated content to align more closely with the input context. This approach is particularly vital for descriptive tasks, such as generating image captions. We demonstrate this in~\cref{fig:applications}. Our method can generate more accurate and detailed descriptions. Moreover, when the generated description is fed into an image generation model like DALLE-3~\cite{betker2023improving}, our method demonstrates its advantages in facilitating image-text alignment.

\vspace{-2mm}
\subsection{Quantitative Evaluation}\label{sec:quant}
\vspace{-2mm}
\paragraph{General VLM benchmarks.} In~\cref{tab:quant_eval}, we evaluate our method on common comprehensive Vision-Language benchmarks, MME~\cite{fu2023mme} and MMBench~\cite{liu2023mmbench}. Prompt Highlighter on LLaVA-v1.5 demonstrates a consistent performance improvement compared to well-trained models by designating the entire image as the highlighted part in the input context. Notably, though these benchmarks primarily assess overall performance with single-token generation and are not designed for user interactions, we still get a competitive place in both MMBench and MME perception. Additionally, further benchmarking with different hyper-parameter selection validates our performance enhancements with LLaVA-v1.5 13B, as evidenced by gains on
POPE{\footnotesize~(85.9 - \textbf{87.8}, \textcolor{purple}{+1.9})}, 
MMB$^{\text{CN}}$ {\footnotesize(63.4 - \textbf{64.0}, \textcolor{purple}{+0.6})}, 
and SQA$^{\text{I}}$ {\footnotesize(71.6 - \textbf{72.4}, \textcolor{purple}{+0.8})}. With LLaVA-NeXT-34B~\cite{liu2024llavanext}, we achieve a SOTA performance on MMB$^{\text{test}}$ with 81.3.

\begin{table}
	\centering
        \scriptsize
	\renewcommand\arraystretch{1.1}
	{
		\begin{tabular}{y{75}|x{30}x{22}x{22}x{22}}
			\toprule
 			method & MME & MMB$_\texttt{dev}$ & MMB$_\texttt{test}$ & PoPE \\
            \midrule
            QWen-VL-Chat~\cite{bai2023qwen} & 1487.5 & 60.6 & 61.8 & - \\
            mPLUG-Owl-2~\cite{ye2023mplug} & 1540.2 & 64.5 & 66.0 & - \\
            \midrule
            LLaVA-v1.5~\cite{liu2023improved} & 1531.3  & 67.7  & 67.0 & 85.9\\
		\cellcolor{Gray}Prompt Highlighter & \cellcolor{Gray}\textbf{1552.5}$_\text{\textcolor{purple}{~+21}}$ & \cellcolor{Gray}\textbf{69.7}$_\text{\textcolor{purple}{~+2.0}}$ & \cellcolor{Gray}\textbf{69.5}$_\text{\textcolor{purple}{~+2.5}}$ & \cellcolor{Gray}\textbf{87.8}$_\text{\textcolor{purple}{~+1.9}}$\\
		\bottomrule
		\end{tabular}}
        \vspace{-2mm}
        \caption{Evaluations on comprehensive VLM benchmarks, including MME Perception~\cite{fu2023mme} and MMBench (MMB)~\cite{liu2023mmbench}.}
        \label{tab:quant_eval}
        \vspace{-3mm}
\end{table}

 \begin{table}
	\centering
        \scriptsize
	\renewcommand\arraystretch{1.1}
	{
		\begin{tabular}{y{64}|x{27}x{18}x{18}x{18}x{18}}
			\toprule
   \multicolumn{1}{l}{}&\multicolumn{1}{c}{}&\multicolumn{2}{c}{text $\rightarrow$ image}&\multicolumn{2}{c}{image $\rightarrow$ text}\\\cmidrule(lr){3-4} \cmidrule(lr){5-6} 
 			method & S-CLIP & R@1 & R@5 & R@1 & R@5\\
			\midrule
                FuseCap~\cite{rotstein2023fusecap} & 0.785 & 37.2 & 62.3 & 47.7 & 72.3\\
                CoCa-CFG~\cite{kornblith2023guiding} & 
                0.808 & 44.6 & \textbf{71.7} & - & - \\
                \midrule
                LLaVA-v1.5~\cite{liu2023improved} & 
                0.809 & 36.6 & 62.2 & 50.7 & 76.4\\
                \cellcolor{Gray}Prompt Highlighter & 
                \cellcolor{Gray}\textbf{0.829} & \cellcolor{Gray}\textbf{45.2} & \cellcolor{Gray}\textbf{71.7} & \cellcolor{Gray}\textbf{62.2} & \cellcolor{Gray}\textbf{85.0}\\
			\bottomrule
		\end{tabular}}
        \vspace{-2mm}
        \caption{CLIP-based reference-free image caption evaluation conducted on the MSCOCO~\cite{lin2014microsoft}.}
        \vspace{-6mm}
        \label{tab:clipscore}
\end{table}

\paragraph{Reliable description.} We then evaluate the reliability of image captions generated by our method. In this case, we utilized the reference-free metric CLIP Score~\cite{hessel2021clipscore}, expressed as $\text{S-CLIP} = 2.5 \cdot \operatorname{max}(\operatorname{cos}(\boldsymbol{v}, \boldsymbol{c}), 0)$, to evaluate the embedding similarity between the image $\boldsymbol{v}$ and the generated caption $\boldsymbol{c}$. We also report the text $\leftrightarrow$ image retrieval recall R@1, R@5 in the CLIP embedding space. As demonstrated in~\cref{tab:clipscore}, on the MSCOCO Karpathy test set~\cite{lin2014microsoft}, our method shows a state-of-the-art CLIP score when compared to recent competitive caption-focused methods~\cite{kornblith2023guiding, rotstein2023fusecap}. Our results exhibit significant advantages across all metrics compared to the baseline.

\paragraph{User study.} We conduct a user study to assess the usability and effectiveness of our method. This study asks participants to rank generation results across five tasks (image captioning, image/text partial highlight generation, text for image generation, and image understanding). Compared with the original inference baselines, the collected 255 valid preference results indicated that \textbf{77.3\%} of users found Prompt Highlighter to generate more correlated results and be beneficial in accomplishing the task objectives. 

More details about the quantitative evaluation can be found in~\cref{sec:supp_eval}.

\vspace{-1mm}
\subsection{Ablation Study}\label{sec:ablation}
\vspace{-2mm}

\paragraph{Module-wise ablation.} In~\cref{tab:module_ablation}, we systematically conduct an ablation study by removing each module of Prompt Highlighter and noting the performance change in MME and MMBench-\texttt{dev}. The results revealed that removing the attention activation module led to the most considerable performance reduction, and combining the highlight guidance and attention activation significantly improves the overall performance. 

\paragraph{Hyper-parameters.}\label{par:hyper_ablation} In~\cref{tab:hyper_ablation}, we explore the impact of three scaling parameters $(\alpha, \beta, \gamma)$ in~\cref{eq:cfg_hl_prob,eq:attntion_add,eq:cfg_embed} for highlight control on the performance of our method. We observed a trade-off between higher concentration (higher $\beta$ and $\gamma$) and general vision-language understanding ability. 

\begin{table}
	\centering
        \scriptsize
	\renewcommand\arraystretch{1.1}
	{
		\begin{tabular}{y{65}|x{23}x{30}|x{22}x{30}}
			\toprule
 			settings & Guidance & Attention & MME & MMB$_\texttt{dev}$\\
			\midrule
			baseline~(our impl.) &  &  & 1528.7 & 67.7\\
                w/ Guidance  & \ding{51}&   & 1531.1 & 68.5\\
                w/ Attention &  & \ding{51} & \underline{1537.2} & \underline{69.5}\\
                \cellcolor{Gray}Full pipeline  & 
                \cellcolor{Gray}\ding{51}& \cellcolor{Gray}\ding{51}& \cellcolor{Gray}\textbf{1552.5} & \cellcolor{Gray}\textbf{69.7}\\
			\bottomrule
		\end{tabular}}
        \vspace{-2mm}
        \caption{A module-wise ablation study.}
        \vspace{-3mm}
        \label{tab:module_ablation}
\end{table}

\begin{table}
	\centering
        \scriptsize
	\renewcommand\arraystretch{1.1}
	{
		\begin{tabular}{y{53}|x{30}x{30}x{30}x{30}}
			\toprule
 			\cellcolor{Gray}$\alpha$~(in~\cref{eq:cfg_embed}) & \cellcolor{Gray}0.0 & \cellcolor{Gray}0.01 & \cellcolor{Gray}0.1 & \cellcolor{Gray}1.0\\
			($\alpha$, 2.0, 1.3) & 1517.6 & \textbf{1552.5} & 1522.3  & \underline{1524.2}\\
                \midrule
                \cellcolor{Gray}$\beta$~(in~\cref{eq:attntion_add}) & \cellcolor{Gray}1.0 & \cellcolor{Gray}2.0 & \cellcolor{Gray}3.0 & \cellcolor{Gray}4.0\\
			(0.01, $\beta$, 1.3) & 1527.2 & \textbf{1552.5} & \underline{1537.3} & 1535.7\\
                \midrule
                \cellcolor{Gray}$\gamma$~(in~\cref{eq:cfg_hl_prob}) & \cellcolor{Gray}1.0 & \cellcolor{Gray}1.3 & \cellcolor{Gray}1.5 & \cellcolor{Gray}2.0\\
			(0.01, 2.0, $\gamma$) & \underline{1537.2} & \textbf{1552.5} & 1532.6  & 1524.0\\
			\bottomrule
		\end{tabular}}
        \vspace{-1mm}
        \caption{Hyper-parameter ablation on MME. We probe for the most suitable value for one of the combinations in $(\alpha, \beta, \gamma)$.}
        \vspace{-1mm}
        \label{tab:hyper_ablation}
\end{table}

\begin{table}[t]
	\centering
        \scriptsize
	\renewcommand\arraystretch{1.1}
	{
		\begin{tabular}{y{65}|x{30}x{55}x{35}}
			\toprule
 			method &  token/s~\textcolor{purple}{$\uparrow$}  & memory~(MB)~\textcolor{purple}{$\downarrow$} & S-CLIP~\textcolor{purple}{$\uparrow$}\\
            \midrule
            Baseline & 6.67  & 16231 & 0.809\\
            Baseline beam=2 & 5.97 & 18537 & 0.807\\
		\cellcolor{Gray}Prompt Highlighter & \cellcolor{Gray} 5.95 & \cellcolor{Gray}17373 & \cellcolor{Gray}0.829\\
		\bottomrule
		\end{tabular}}
        \vspace{-2mm}
        \caption{Evaluation on inference speed and GPU memory. The memory is dominated by the model weight, consuming 13971 MB.}
        \label{tab:speed_mem}
        \vspace{-5mm}
\end{table}


\vspace{-1mm}
\subsection{Discussions}\label{sec:interp}
\begin{figure}[t]
    \centering
     \includegraphics[width=0.9\linewidth]{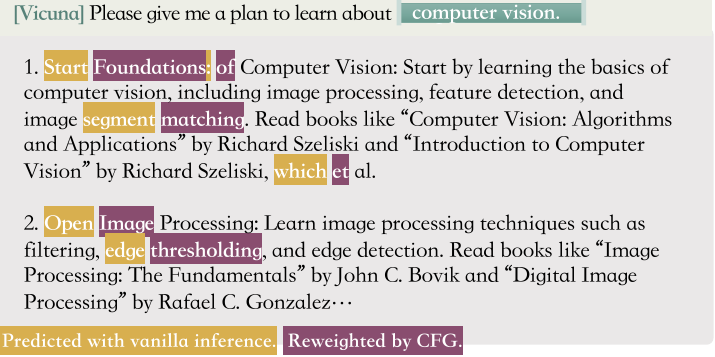}
     \vspace{-2mm}
     \caption{An example of token changed with CFG.}\label{fig:inter_cfg}
\end{figure}

\vspace{-1mm}
\paragraph{Prediction control by CFG.} Given that our highlight guidance operates on the predicted token probability, similar to LLM-CFG~\cite{sanchez2023stay}, we further investigate the semantic-level distinctions in normal and unconditional branches. This is visualized through an example of token prediction in~\cref{fig:inter_cfg}. When the embeddings associated with the highlighted tokens are perturbed, the unconditional-context branch tends to predict unrelated to the highlighted parts. This, in turn, enables the rectification of responses in text-generation tasks.

\begin{figure}[t!]
    \centering
     \includegraphics[width=0.9\linewidth]{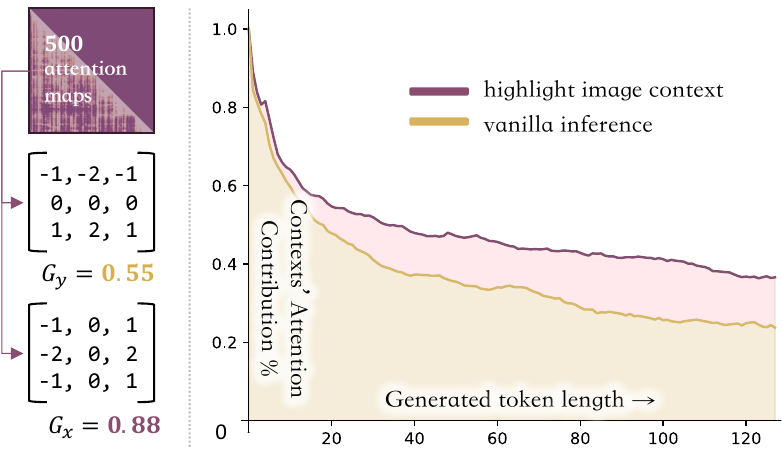}
     \vspace{-2mm}
     \caption{\textit{\textbf{Left}}: A simple verification of the vertical band-like pattern in the attention map, with a report on the gradient summation. \textit{\textbf{Right}}: Following~\cite{zhang2023realworld}, we present a visualization displaying the average contribution of context's attention during generation.}\label{fig:inter_attn}
     \vspace{-2mm}
\end{figure}

\paragraph{Attention activation.} To confirm that the attention activation operates as anticipated, we not only visualize the activated attention scores corresponding to different regions in~\cref{fig:qformer_vis}, but we also demonstrate its effectiveness using 500 attention maps from the caption experiment in~\cref{fig:inter_attn}. Firstly, we validate the band-like pattern property discussed in~\cref{sec:attention_activation}, by comparing the vertical and horizontal gradients. We observed a significant gradient gap with $G_x > G_y$ in all 500 cases. Given this property, the attention activation can capture a higher attention probability $p$, as defined in \cref{eq:attntion_explain}, from the given contexts. This attention contribution plotted in~\cref{fig:inter_attn} is denoted as $\sum_{m_i=1}(p_i)/{\sum_{j<i}(p_j)}$. Consequently, attention activation leads to results that adhere more closely to the input context.

\paragraph{Limitations and future work.}\label{sec:limitations} While our approach introduces a novel method for controlling generation in multi-modal LLMs, it has certain limitations:
\textit{(a). Additional computations:} Our method requires an extra decoding branch, which brings additional computational overhead and GPU memory requirements. However, these additional loads are marginal and acceptable with the batched inference. We validate this by the caption experiment in~\cref{tab:speed_mem}.
\textit{(b). Dependence on base model:} Content generation quality is tied to the base model's capabilities, which may result in over-emphasis or miss-emphasis on highlighted parts when using poorly-trained base models.

One direction for future work will be to create a more intuitive highlighting scheme. We also aim to extend our method to support a greater variety of interactions.

\vspace{-2mm}
\section{Conclusion}\label{sec:conclusion}
\vspace{-2mm}
We introduce Prompt Highlighter, a novel paradigm for user-model interactions in multi-modal LLMs, offering output control through a token-level highlighting mechanism. This approach, requiring no extra training, competes well on standard benchmarks and provides reliable generation outputs by merely highlighting input context. Further, diverse applications demonstrate its intuitive usability and effectiveness in enhancing control over the generation process. This work represents a promising direction for enhancing user control in multi-modal LLMs, and we anticipate it will inspire further research.

{
\paragraph{Acknowlegdement} This work was supported in part by the Research Grants Council under the Areas of Excellence scheme grant AoE/E-601/22-R and the Shenzhen Science and Technology Program under No.~KQTD20210811090149095.
}


\clearpage
\etoctoccontentsline{part}{Supplementary Material}

\appendix

\setlength{\cftbeforesecskip}{0.5em}
\cftsetindents{section}{0em}{1.8em}
\cftsetindents{subsection}{1em}{2.5em}
\cftsetindents{subsubsection}{3.0em}{3.5em}

\renewcommand{\contentsname}{Appendix Contents}
\hypersetup{linkbordercolor=black,linkcolor=black}
\localtableofcontents
\hypersetup{linkbordercolor=red,linkcolor=red}

\section{Experiment Details}\label{sec:exp_detail}

\subsection{Workflow of the Prompt Highlighter}

For a more comprehensive understanding of the method part of Prompt Highlighter, we provide pseudo-code algorithmic workflows in~\cref{alg:prompt_highlighter} and~\cref{alg:attention_activation}. In this context, the attention activation outlined in~\cref{alg:attention_activation} acts as the modified forward function in all Self-Attention layers during the multi-modal LLMs' inference $\tilde{\mathbf{P}}_{\Theta}(x_\eta | \mathbf{s}, \mathbf{m})$, as discussed in~\cref{sec:attention_activation} and illustrated in~\cref{alg:prompt_highlighter}.

\begin{algorithm}[ht]
\caption{Highlighted Guidance Control}
\label{alg:prompt_highlighter}
\begin{algorithmic}[1]
    \small
    \REQUIRE Pre-trained LLM or VLM decoder $\mathbf{P}_{\Theta}$, input multi-modal tokens $\mathbf{x}$, binary highlight mask $\mathbf{m}$, guidance strength $\gamma$, and scaling factor $\alpha, \delta$
    
    \STATE Initialize token sequence $\mathbf{x} = \{x_1,\ldots,x_N\}$
    \STATE Initialize conditional embedding $\mathbf{s} = f(\mathbf{x})$ and unconditional embedding $\bar{\mathbf{s}}$ as $\bar{\mathbf{s}} = (\alpha - 1)\mathbf{m} \cdot f(\mathbf{x}) + f(\mathbf{x})$~(\cref{eq:cfg_embed}).
    \STATE Initialize current token index $\eta = N$, output token sequence $\mathbf{y} = \text{[ ]}$
    \newline
    \WHILE{$x_\eta$ is not \texttt{</s>}}
        \STATE Calculate the greedy decoded prediction $x_\eta$ in~\cref{eq:cfg_hl_prob}:
        \STATE { $\begin{multlined}x_{\eta+1} = \operatorname{argmax}(\gamma\log\bar{\mathbf{P}}_{\Theta}(x_{\eta+1} | \mathbf{s}, \mathbf{m})\\
        - (\gamma - 1)\log\bar{\mathbf{P}}_{\Theta}(x_{\eta+1} | \mathbf{\bar{s}}, -\delta\mathbf{m}))\end{multlined}$ }
    
        \STATE Update output sequence:
        \STATE $\mathbf{y}\text{.append}(\mathbf{P}_{\Theta}\text{.tokenizer.decode}(x_{\eta+1}))$
        \STATE Update next token embedding and mask: 
        \STATE $s_{\eta+1} = \bar{s}_{\eta+1} = f(x_{\eta+1}), m_{\eta+1} = 0, \eta = \eta+1$
    \ENDWHILE
    \RETURN Output sequence $\mathbf{y}$
\end{algorithmic}
\end{algorithm}

\begin{algorithm}[ht]
\caption{Attention Activation in Self-Attention Layer. For simplicity, we demonstrate with single-head attention.}
\label{alg:attention_activation}
\begin{algorithmic}[1]
    \small
    \REQUIRE Input hidden state $\mathbf{q}=\{q_1, \ldots, q_{\eta-1}\}$ and vanilla attention mask $\mathbf{m}_{\text{attn}}$, highlight mask $\mathbf{m}$, Scaling factor $\beta$, Self-Attention Module $M_\text{SA} \subset \tilde{\mathbf{P}}_\Theta$.
    \newline
    \STATE $(Q, K, V) = M_\text{SA}$.proj\_qkv($\mathbf{q}$)
    \STATE Calculate the attention score map:
    \STATE $\mathbf{k} = Q @ K^\top + \mathbf{m}_{\text{attn}}$
    \STATE Initialized the highlighted attention mask:
    \STATE $\mathbf{w} = \mathbf{m}$.expand\_as($\mathbf{k}$)
    \STATE Activate score via attention map~(\cref{eq:attntion_add,eq:qformer}):
    \STATE $\mathbf{h} = \log(\beta)\mathbf{w} + \mathbf{k}$
    \STATE Then complete the remaining operations in self-attention:
    \STATE $\mathbf{p} = \operatorname{softmax}(\mathbf{h})$
    \RETURN $\mathbf{q}_{\text{out}} = M_\text{SA}$.proj\_out($\mathbf{p} @ V$)
\end{algorithmic}
\end{algorithm}

\subsection{Quantitative Evaluation}\label{sec:supp_eval}
\subsubsection{VLM Benchmarks} \vspace{-1mm}
We use the same prompts and test scripts for our benchmark tests as those in the LLaVA-v1.5 codebase~\cite{liu2023improved}. The only difference is our highlighting of the image context in the inputs~(\ie, $m_i=1$ when $s_i\in\mathbf{s}^{\text{im}}$). We adopt softmax, instead of log softmax used in the multiple-token prediction tasks, as the token probability rescale function. As discussed in~\cref{sec:ablation}, we observe slight variations in results across different test sets based on the parameters of attention activation. In~\cref{tab:supp_quant_eval}, we showcase the impact of changes in the $\beta$ value on the benchmark tests. Our method still achieves the second rank on the current MMBench-$\texttt{dev}$ leaderboard, even when using a uniform parameter of $\beta=2.0$.

\begin{table*}[ht!]
	\centering
        \footnotesize
	\renewcommand\arraystretch{1.1}
	{
		\begin{tabular}{l|ccccccc|ccccccc}
			\toprule
                \multicolumn{1}{l}{}&\multicolumn{7}{c}{MMBench-\texttt{dev}}&\multicolumn{7}{c}{MMBench-\texttt{test}}\\\cmidrule(lr){2-8} \cmidrule(lr){9-15} 
 			method & \textbf{Overall} & LR & AR & RR & FP-S & FP-C & CP & \textbf{Overall} & LR & AR & RR & FP-S & FP-C & CP\\
            \midrule
            MMICL~\cite{zhao2023mmicl}& \underline{67.9} & \textbf{49.2} & \underline{71.6} & \textbf{73.0} & 66.7 & 57.2 & 77.2  & 65.2 & \textbf{44.3}	& \textbf{77.9} & \textbf{64.8} & 66.5 & 53.6 & 70.6 \\
            mPLUG-Owl2~\cite{ye2023mplug} & 66.5 & 32.2 & \textbf{72.4} & 60.9 & 68.6 & 60.1 & 79.4  & 66.0 & \underline{43.4} & 76.0 & 62.1 & 68.6 & 55.9 & 73.0 \\
            Sphinx~\cite{lin2023sphinx} & 67.2  & 33.1 & 67.3 & 58.3 & \textbf{74.4} & \textbf{59.4} & \underline{80.7}  & \underline{67.5} & 32.9	& 73.6& 57.8& \underline{72.1}& \textbf{63.2} & \textbf{79.2}\\
            \midrule
            LLaVA-v1.5~\cite{liu2023improved} & 67.7  & 41.7 & 69.7 & 63.5 & 70.0 & \underline{59.3} & 80.2  &67.0& 39.9 & 74.7& 61.6& 70.9& 59.9& 75.4\\
          \cellcolor{Gray}+ Prompt Highlighter  & \cellcolor{Gray}\textbf{69.7} & \cellcolor{Gray}\underline{44.2}& \cellcolor{Gray} 70.6& \cellcolor{Gray}\underline{68.7} & \cellcolor{Gray}\underline{73.7} & \cellcolor{Gray}\underline{59.3} & \cellcolor{Gray} \textbf{80.9}  & \cellcolor{Gray} \cellcolor{Gray}\textbf{69.5}& \cellcolor{Gray}42.6 & \cellcolor{Gray}\underline{77.5} & \cellcolor{Gray}\underline{64.3} & \cellcolor{Gray}\textbf{75.0} & \cellcolor{Gray}\underline{62.0} & \cellcolor{Gray}\underline{76.4} \\
	   \cellcolor{Gray}Improvement & \cellcolor{Gray}\textcolor{purple}{+2.0}  & \cellcolor{Gray}\textcolor{purple}{+2.5} & \cellcolor{Gray}\textcolor{purple}{+0.9} & \cellcolor{Gray}\textcolor{purple}{+5.2} & \cellcolor{Gray}\textcolor{purple}{+3.7} & \cellcolor{Gray}\textcolor{purple}{0.0} & \cellcolor{Gray}\textcolor{purple}{+0.7}  & \cellcolor{Gray}\textcolor{purple}{+2.5} & \cellcolor{Gray}\textcolor{purple}{+2.7}	& \cellcolor{Gray}\textcolor{purple}{+2.8} & \cellcolor{Gray}\textcolor{purple}{+2.7} & \cellcolor{Gray}\textcolor{purple}{+4.1} & \cellcolor{Gray}\textcolor{purple}{+2.1} & \cellcolor{Gray}\textcolor{purple}{+1.0} \\
		\bottomrule
		\end{tabular}}
        \caption{Detailed comparison in MMBench-\texttt{dev} and MMBench-\texttt{test}\cite{liu2023mmbench}. The categories include Logic Reasoning~(LR), Attribute Reasoning~(AR), Relation Reasoning~(RR), Instance-Level Fine-Grained Perception~(FP-S), Cross-Instance Fine-Grained Perception~(FP-C), and Coarse Perception~(CP). The improvement of our method over the LLaVA-v1.5~\cite{liu2023improved} baseline is reported in the last row for each category. We highlight \textbf{the best} and underline \underline{the second best} result for each column.}
        \label{tab:supp_quant_detail}
\end{table*}

\begin{table}[t]
	\centering
        \footnotesize
	\renewcommand\arraystretch{1.1}
	{
		\begin{tabular}{y{75}|x{37}x{37}x{37}}
			\toprule
 			method & MME & MMB$_\texttt{dev}$ & MMB$_\texttt{test}$ \\
            \midrule
            LLaVA-v1.5~\cite{liu2023improved} & 1531.3  & 67.7  & 67.0\\
            \midrule
          PH (0.01, \textbf{3.0}, 1.3) & 1537.3  & \textbf{69.7}  & 68.7\\
		 PH (0.01, \textbf{2.0}, 1.3) & \textbf{1552.5} & 68.7 & \textbf{69.5}\\
            \midrule
           PH (submitted) & \textbf{1552.5} &  \textbf{70.1} & \textbf{70.7}\\
		\bottomrule
		\end{tabular}}
        \caption{Hyper-parameter settings in VLM benchmarks. Evaluations are conducted on MME Perception~\cite{fu2023mme} and MMBench ~(MMB)~\cite{liu2023mmbench}. We list a combination of hyper-parameters $(\alpha, \beta, \gamma)$ for our method~(PH for Prompt Highlighter).}
        \label{tab:supp_quant_eval}
\end{table}

We adopt the MME and MMBench datasets because they offer comprehensive evaluations of VLMs across multiple dimensions and feature over 10K of VQA questions. This makes them ideal for assessing the overall competency of Vision-Language Models. We present a detailed evaluation result of MMBench in~\cref{tab:supp_quant_detail}. With the same pre-trained weight as LLaVA-v1.5, our training-free method consistently outperforms the baseline across nearly all evaluation dimensions. Given that our approach is compatible with VLM frameworks that use token-level embedding input and a Transformer-based language decoder, it positions the Prompt Highlighter as a \textbf{training-free plugin} for integrating context-highlighting abilities into these models. 
We further evaluate the MMBench benchmark using the current state-of-the-art model, InternLM-VLComposer~\cite{zhang2023internlmxcomposer}. Our method demonstrated a performance improvement, increasing the SOTA performance from 74.8 to 75.3 in the \texttt{dev} split
with this representative Q-Former-based approach. These results suggest that though our method is designed to create new interactive ways instead of focusing on benchmark improvement, it still proves to be competitive in general applications.



\subsubsection{Reliable Descriptions}
As discussed in~\cref{sec:ablation}, we observe a trade-off between the attention concentration and QA performance that appeared in the benchmarks. Therefore, compared with experiments in VLM benchmarks, we adopt a relatively more aggressive parameter setting in the MSCOCO caption experiment: $(\alpha, \beta, \gamma) = (0.01, 7.0, 2.0)$. 
We utilize a CLIP ViT-Base-32 model~\cite{dosovitskiy2021an, radford2021learning} to test the CLIP Score~\cite{hessel2021clipscore} and to extract embeddings for both text $\rightarrow$ image and image $\rightarrow$ text retrieval metrics. Our experiment uses a simple input prompt, \sethlcolor{Gray}\hl{ \texttt{'Describe this image.'} }. In the comparison in~\cref{tab:clipscore}, we report the official result reported by CoCa-CFG~\cite{kornblith2023guiding} with the same $\gamma$ value of 2.0\footnote{CoCa-CFG achieves caption $\rightarrow$ image retrieval recall with R@1=49.4 and R@5=75.7 with $\gamma=3.0$, while it remains the same S-CLIP=0.808.}. As CLIP has a maximum token length limitation of 77, we split captions at semicolons and periods and input the longest possible concatenated sequence to circumvent this limitation. For all text-to-image generation results in figures, we query the generated text description to the Bing Image Creator powered by DALLE-3~\cite{betker2023improving} {\small[\url{www.bing.com/images/create}]}.

\vspace{-2mm}
\subsubsection{User Study}
We present the detailed results of the user study in~\cref{tab:supp_user_study}. Using different sub-task examples, we engaged various baseline models for different tasks to demonstrate the consistent preference for Prompt Highlighter across a broad array of frameworks and tasks. The user preference results affirm that our method delivers superior performance and flexibility, consistently outperforming the baseline models across a wide range of tasks.

\begin{table}[t]
	\centering
        \footnotesize
	\renewcommand\arraystretch{1.1}
	{
		\begin{tabular}{l|lc}
			\toprule
 			sub-task & baseline &  preference \\
            \midrule
            Text Partial Highlighter & Vicuna-13B &70.6\%\\
            Image Partial Highlighter  & LLaVA & 74.5\%\\
            Image Understanding  & InstructBLIP  & 76.5\%\\
            Image Caption & LLaVA-v1.5 & 80.4\%\\
            Description $\rightarrow$ Image & LLaVA-v1.5 + DALLE-3 &84.3\%\\
            \cellcolor{Gray}\textbf{Overall}& \cellcolor{Gray}- &\cellcolor{Gray}\textbf{77.3\%}\\
		\bottomrule
		\end{tabular}}
        \caption{A fine-grained user study result.}
        \label{tab:supp_user_study}
\end{table}

\begin{figure*}[t!]
    \centering
     \includegraphics[width=1\linewidth]{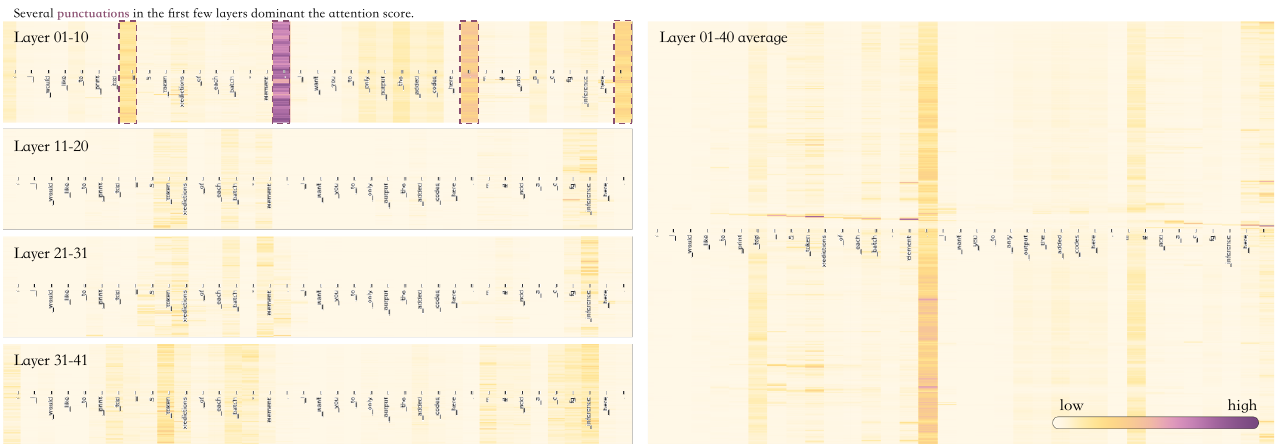}
     \vspace{-2mm}
     \caption{An additional example of attention maps across different layer segments~(\textit{\textbf{left}}) and the averaged visualization with more generated tokens~(\textit{\textbf{right}}). We mark tokens representing punctuation, which draw significant attention in the initial layers. Beyond these tokens, a consistent correlation between tokens forms a band-like pattern that can be observed across all layers.}\label{fig:supp_attn_layerwise}
     \vspace{-2mm}
\end{figure*}

\subsection{Attention Map Visualization}

In this section, we delve into the attention map visualization, expanding on the verification experiments discussed in~\cref{sec:interp}. In~\cref{fig:supp_attn_layerwise}, we illustrate an example akin to~\cref{fig:attn_vis}, demonstrating the persistence of the band-like pattern in different layer segments in the Vicuna-13B model~\cite{chiang2023vicuna}. The band-like property is clear and consistently evident across various layers. Given that these band-like token activation patterns are distributed across different layers, we simply incorporate the attention activation modification in all self-attention layers. Furthermore, the attention maps referenced in~\cref{fig:inter_attn} to calculate gradients and attention contributions are computed based on the averaged attention map across layers and heads.

\begin{figure}[h]
    \centering
     \includegraphics[width=1\linewidth]{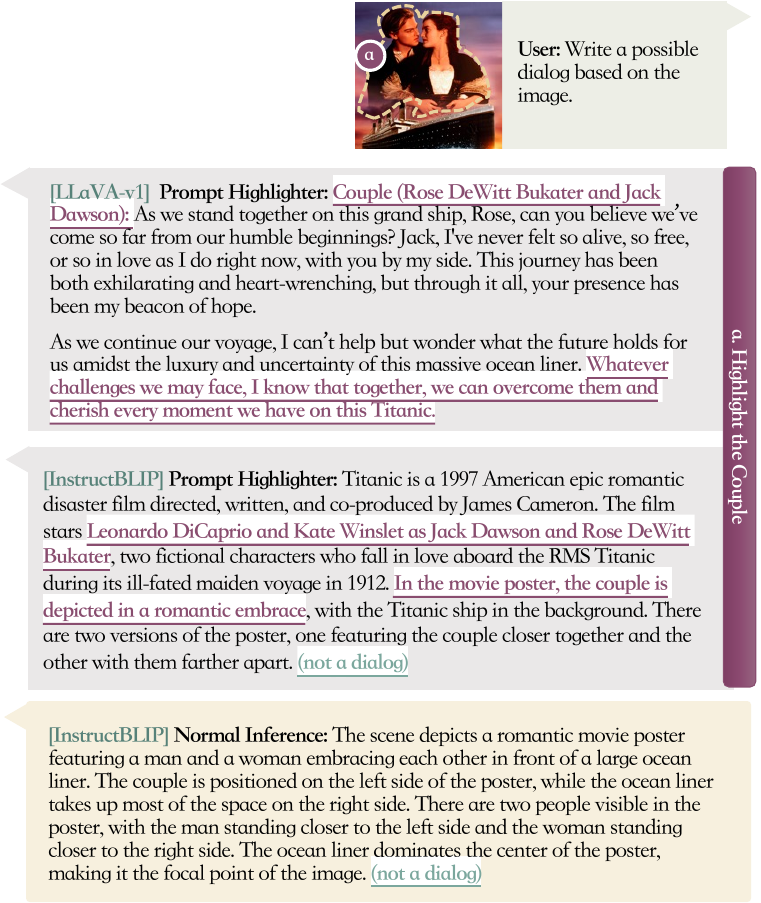}
     \caption{An example illustrating the influence of different base models. Given the same highlighted context, the output of the model may be constrained by the capabilities of the base model.}\label{fig:supp_basemodel}
\end{figure}

\subsection{Limitation Analysis}\label{sec:supp_limitations}
\subsubsection{Speed and Memory}

\begin{figure}
    \centering
    \includegraphics[width=1.0\linewidth]{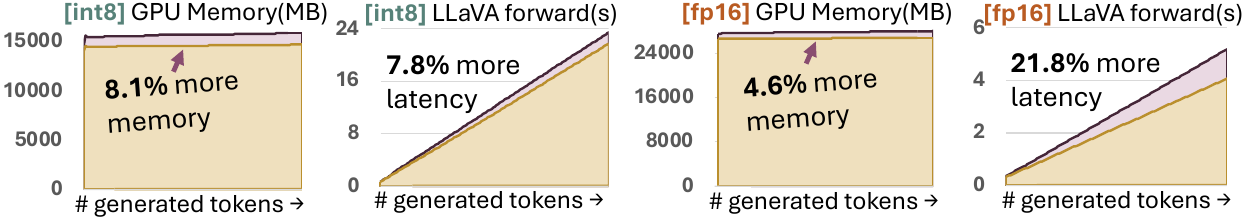}
    \vspace{-3mm}
    \caption{Speed and memory overhead analysis. We evaluate VLM inference both in int8 and fp16 precision.}\label{fig:speedmem_curve}
     \vspace{-3mm}
\end{figure}

For processing two tokens compared to single-beam (greedy) inference in image description, batched inference enables parallel computation, adding only 7\% and 10\% to the memory and speed costs, respectively.
We corroborate this with plotted relation curves for processing $N=0,\dots,140$ tokens, demonstrating that our method maintains controllability with minimal extra costs.

As discussed in~\cref{sec:interp}, Prompt Highlighter imposes an additional computational load on the unconditional branch, and we validate its tangible impact in~\cref{tab:speed_mem}. Memory consumption with 8-bit loading confirms that the Prompt Highlighter is designed to be memory-efficient, making it feasible to run inference on a 13B model using commonly available GPUs, such as the NVIDIA-3090. This added load becomes noticeable when evaluating single-token prediction tasks, as the model is required to compute double KV values among tokens in its initial inference step. However, when employing KV-cache in the multiple-token generation, this extra load can be alleviated, as the doubled calculation with cached KVs does not become the bottleneck for inference speed or GPU memory consumption. 

For processing two tokens compared to single-beam (greedy) inference in image description, batched inference enables parallel computation, adding only 7\% and 10\% to the memory and speed costs, respectively. We corroborate this with plotted relation curves in~\cref{fig:speedmem_curve} for processing $N=0,\dots,140$ tokens, demonstrating that our method maintains controllability with minimal extra costs.
We also confirm that the commonly utilized inference strategy of introducing an extra beam search branch does not impair the quality of the descriptive task~(S-CLIP 0.809$\rightarrow$0.807 in~\cref{tab:speed_mem}), despite imposing an additional computational load. This also illustrates the practical effectiveness of Prompt Highlighter.

\subsubsection{Constrained by the Base-Model}
Our method stems from the base model and its plug-and-play nature. As illustrated in~\cref{fig:supp_basemodel}, the base model InstructBLIP~\cite{dai2023instructblip} is not adept at dialog generation. Consequently, with a highlighted prompt context, despite the output text being more related to the highlighted part, it fails to generate a dialogue effectively. This suggests that while the Prompt Highlighter can augment existing abilities in pre-trained multi-modal LLMs, it cannot endow them with new capabilities.

\section{Additional Disccusions}
\begin{figure}[t]
    \centering
     \includegraphics[width=1\linewidth]{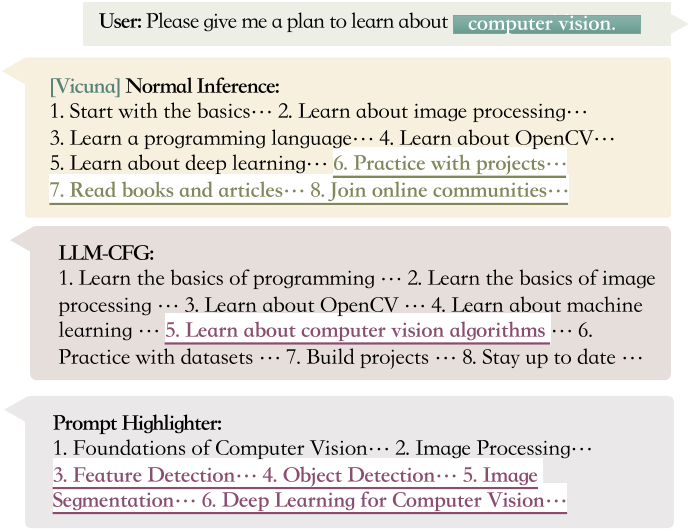}
     \vspace{-2mm}
     \caption{One example of the comparison with LLM-CFG~\cite{sanchez2023stay}.}\label{fig:cfg-llm}
     \vspace{-2mm}
\end{figure}

\begin{figure}[t]
    \centering
     \includegraphics[width=1\linewidth]{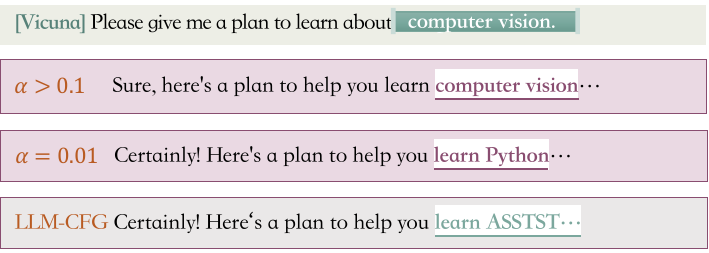}
     \vspace{-2mm}
     \caption{Output example for the unconditional branch with different unconditional context designs.}\label{fig:uncond_out}
     \vspace{-2mm}
\end{figure}

\begin{figure}[t]
    \centering
     \includegraphics[width=1\linewidth]{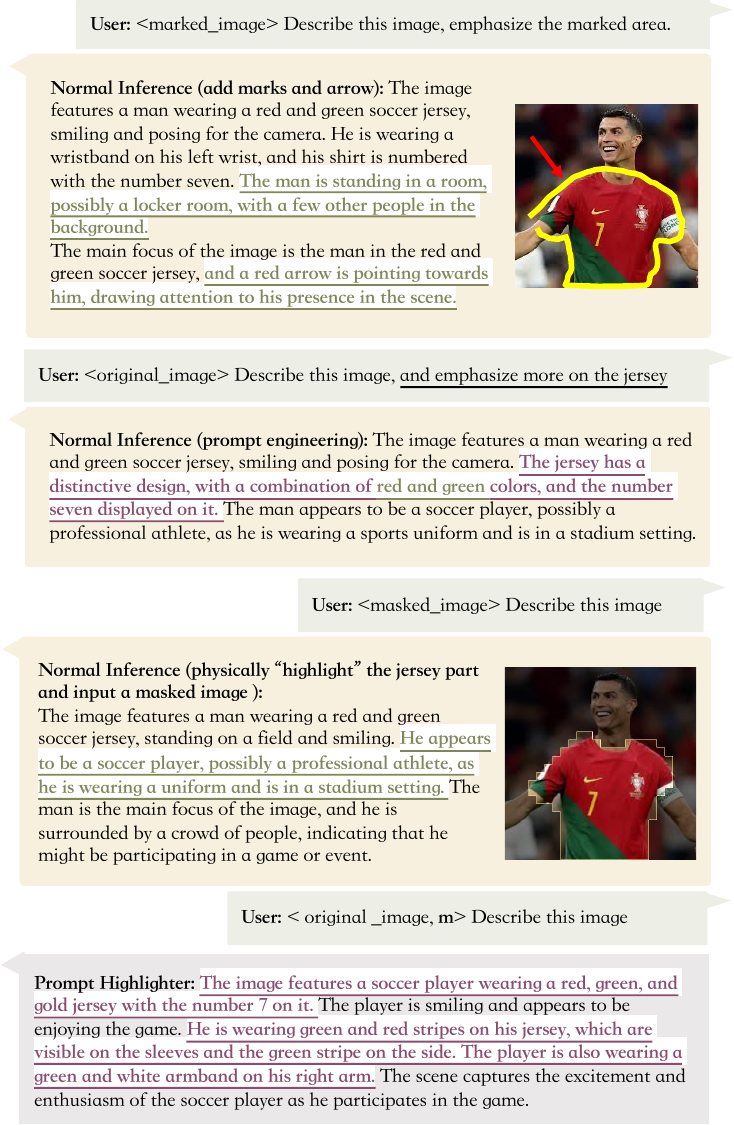}
     \vspace{-2mm}
     \caption{A comparison of Prompt Highlighter with different vision-language prompt manipulation approaches.}\label{fig:supp_hl_compare}
     \vspace{-2mm}
\end{figure}

\begin{figure*}[p]
    \centering
    \vspace{5mm}
     \includegraphics[width=\textwidth,height=\textheight,keepaspectratio]{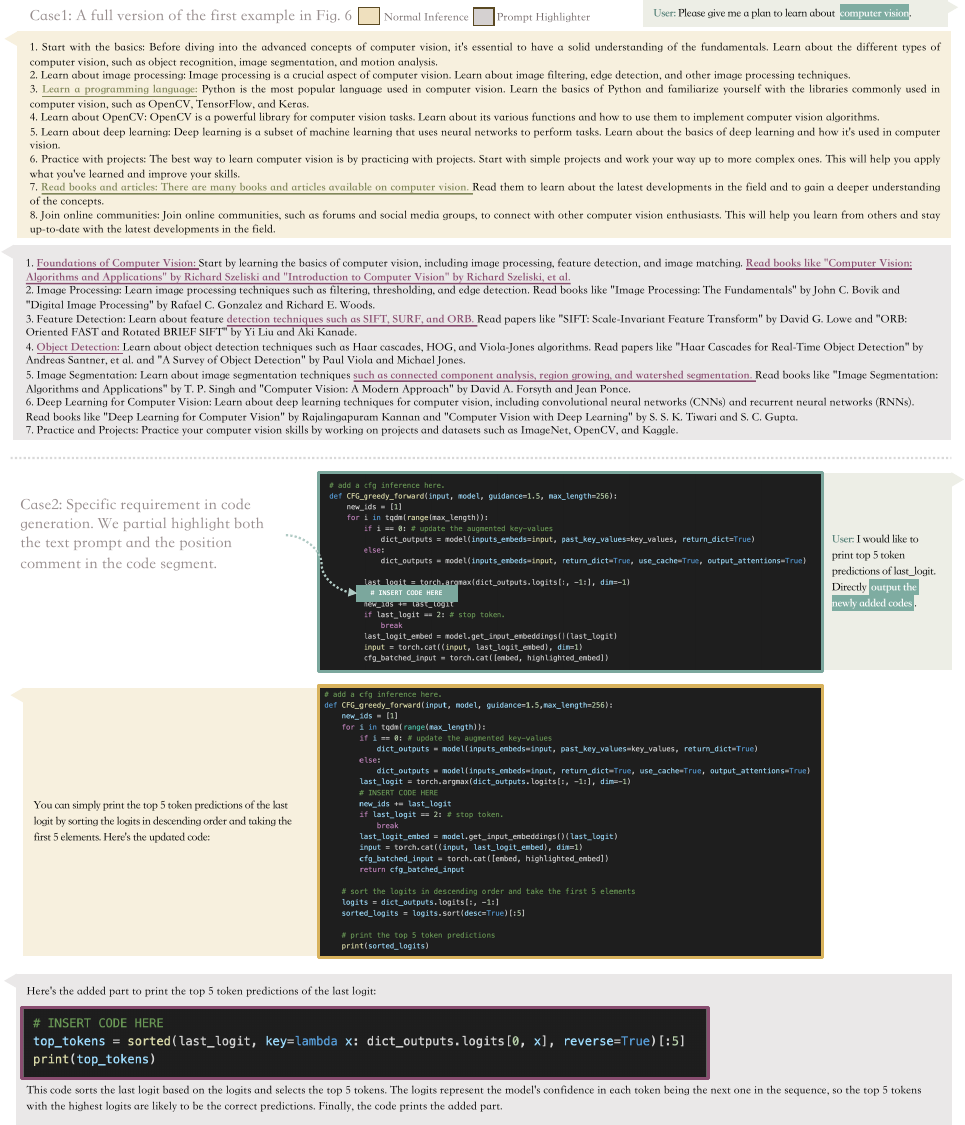}
     \caption{Two more examples in pure text partial highlighter. A Vicuna-13B-v1.1~\cite{chiang2023vicuna} is used as the base model.}\label{fig:supp_txt_hl}
     \vspace{5mm}
\end{figure*}

\begin{figure*}[p]
    \centering
    \vspace{7mm}
     \noindent\makebox[\textwidth]{\includegraphics[width=\textwidth,height=\textheight,keepaspectratio]{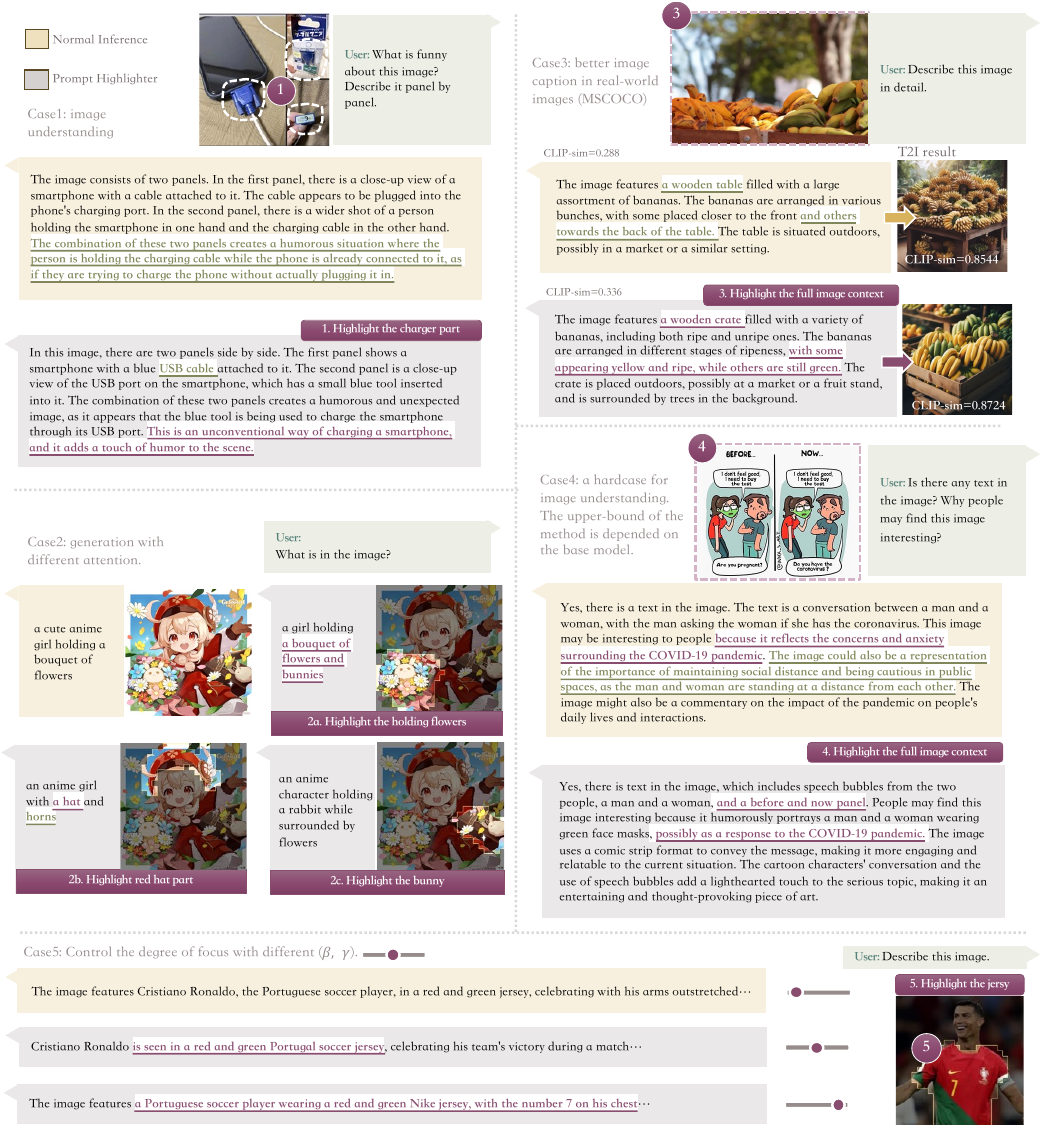}}
     \caption{More results with multi-modal inputs. We demonstrate three applications~(1,2,3,5) and one hard case~(4) for image understanding. We use InstructBLIP-Vicuna-13B~\cite{dai2023instructblip} as the base model of cases 1,2,5 and LLaVA-v1.5~\cite{liu2023improved} 13B as the base model of cases 3,4.}\label{fig:supp_img_hl}
     \vspace{7mm}
\end{figure*}

\begin{figure*}[p]
    \centering
     \includegraphics[width=\textwidth,height=\textheight,keepaspectratio]{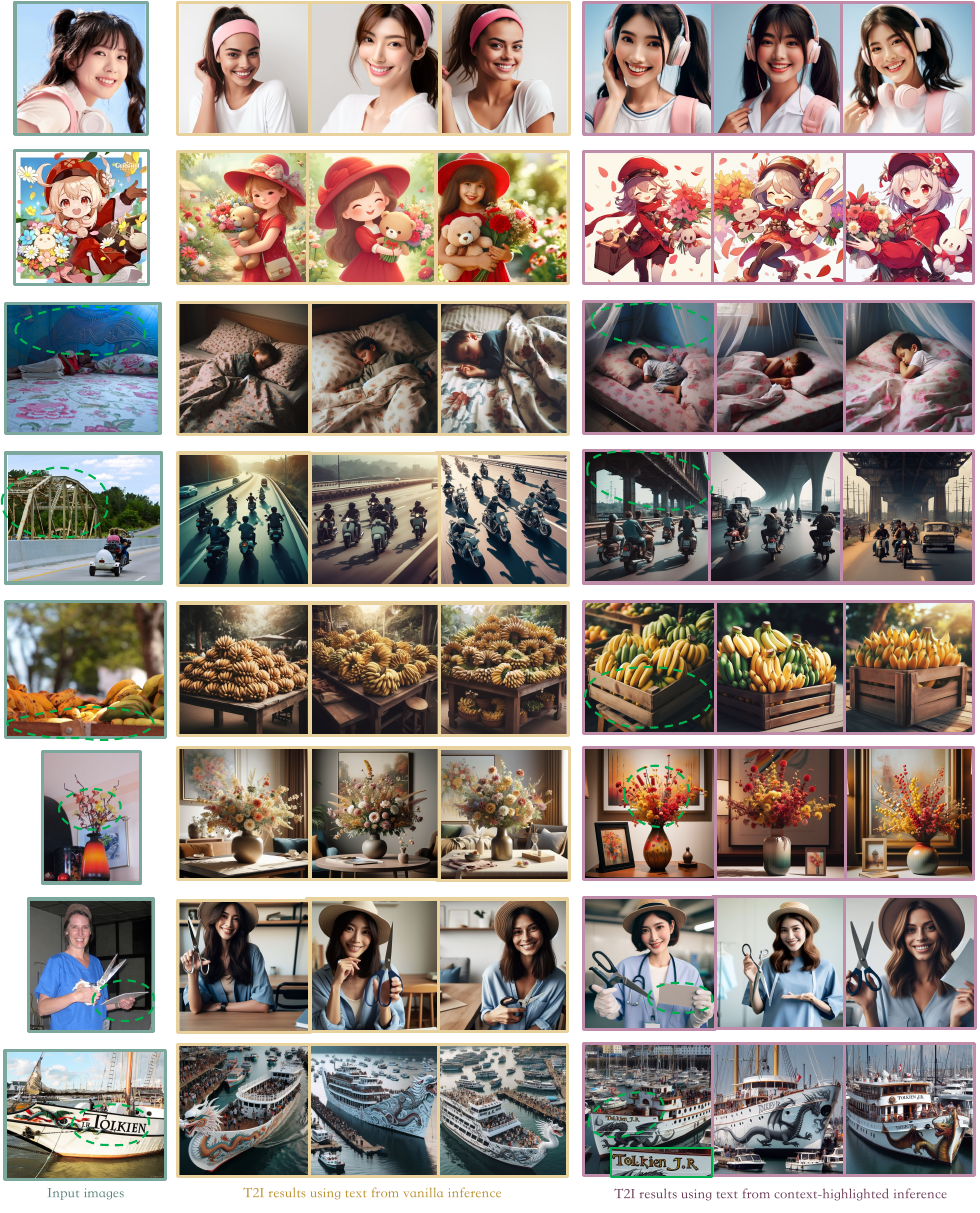}
     \vspace{-5mm}
     \caption{More text-to-image examples with image generation by DALLE-3~\cite{betker2023improving}. We have marked areas in some figures that generate better results due to different captions.}\label{fig:supp_dalle3}
     \vspace{-2mm}
\end{figure*}

\subsection{Compare with LLM-CFG}
The comparison between Prompt Highlighter and LLM-CFG~\cite{sanchez2023stay} is visually demonstrated in~\cref{fig:cfg-llm}. Our method stands out by offering a more granular control mechanism that allows specific tokens within the context to be highlighted. Conversely, LLM-CFG, with its basic prompt-level differentiation and without a mechanism to directly manipulate model feature interactions (\eg, our attention activation strategy), often generates outputs that bear noticeable similarity to the original, regardless of substantial $\gamma$ increases.

Using the identical prompt as in~\cref{fig:cfg-llm}, we consolidate this claim by exhibiting the independent inference outcomes of the unconditional branch in~\cref{fig:uncond_out}. In addition, we demonstrate the significance of assigning the $\alpha$ value at the embedding layer. As shown in~\cref{fig:uncond_out} and as per our discussion in~\cref{sec:highlight_method}, it becomes apparent that a small scalar multiplication $\alpha$ in the embedding space leaves the overall contextual understanding of the unconditional branch largely unaffected. Such a fine-grained difference assists in making more distinguishing token choices during the inference process in~\cref{eq:cfg_hl_prob}. In contrast, LLM-CFG's unconditional branch omits part of the prompt, resulting in an obstacle to generating results that effectively underscore the highlighted sections.

\subsection{Compare with other Highlight Methods}
To demonstrate the non-trivial nature of our method, we evaluated it alongside several intuitive prompt engineering approaches, which include adding adjectives, capitalizing text, and explicitly marking in images. Comparative examples in LLM and VLM are provided in~\cref{fig:teaser_image,fig:supp_hl_compare}, respectively. Our observations suggest that the employment of additional prompts or explicit emphasis can occasionally lead to unpredictable results. This is primarily due to the introduction of extra information, while models may not truly emphasize correlated parts. In contrast, our method seamlessly accommodates all context information and produces focused text outputs. This demonstrates that the design of our method not only offers a high degree of flexibility but also enables the generation of more contextually appropriate outputs.

\subsection{Highlighting the Whole Image.}
We demonstrate in ~\cref{alg:attention_activation} and ~\cref{sec:attention_activation} that self-generated tokens may lead to model hallucinations. The model's attention is steered towards trustworthy content by purposefully highlighting the reliable input context, such as the entire image, mitigating hallucinations in extended text generation. An example from MME is provided below to show how this approach effectively curbs hallucinations during evaluations.

\begin{figure}[H]
\vspace{-3pt}
    \centering
    \includegraphics[width=1.0\linewidth]{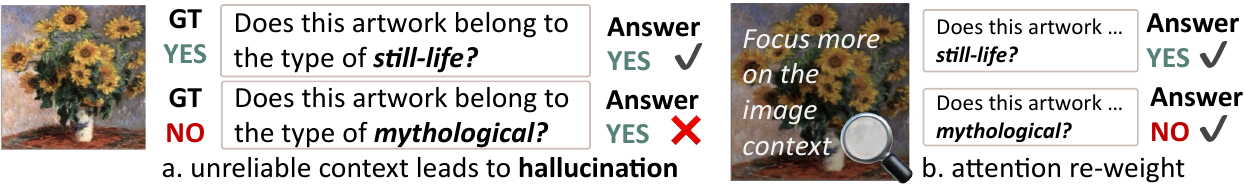}
    \caption{An example of highlighting image context to eliminate hallucination.}
\vspace{-5pt}
\end{figure}

\section{Showcases}
\subsection{More Visual Results}
Additional results of Prompt Highlighter in LLM and VLM are provided in~\cref{fig:supp_txt_hl} and~\cref{fig:supp_img_hl}, respectively.

In the first example of~\cref{fig:supp_txt_hl}, we showcase the primary part of the output of the example in~\cref{fig:showcase}. Our results concentrate more on the highlighted section, offering professional reference books to support user learning ``computer vision''. The second example exhibits a case for a multi-segment mask input, where the model is steered to output code snippets at the appropriate positions by highlighting the requirement part of the text and the comment as the position identifier.
\cref{fig:supp_img_hl} demonstrates a range of applications and capabilities of Prompt Highlighter in VLM from various perspectives, including image content comprehension (case 1), focused generation (case 2), improved image descriptions (case 3), and explicit generation control (case 5). These instances underscore the performance enhancement and customizable generation control facilitated by Prompt Highlighter.
We also present some challenging cases due to the limited perceptual capacity of the base model.

Moreover, we present results of text-to-image generation based on the text descriptions generated by our method in~\cref{fig:supp_dalle3}. These outcomes vouch for the superior capability of our approach in descriptive tasks and its proficiency in generating higher-quality image-text matching data.

\subsection{Multi-Round Interactive Conversation}
\begin{figure*}[ht]
    \centering
     \includegraphics[width=1\linewidth]{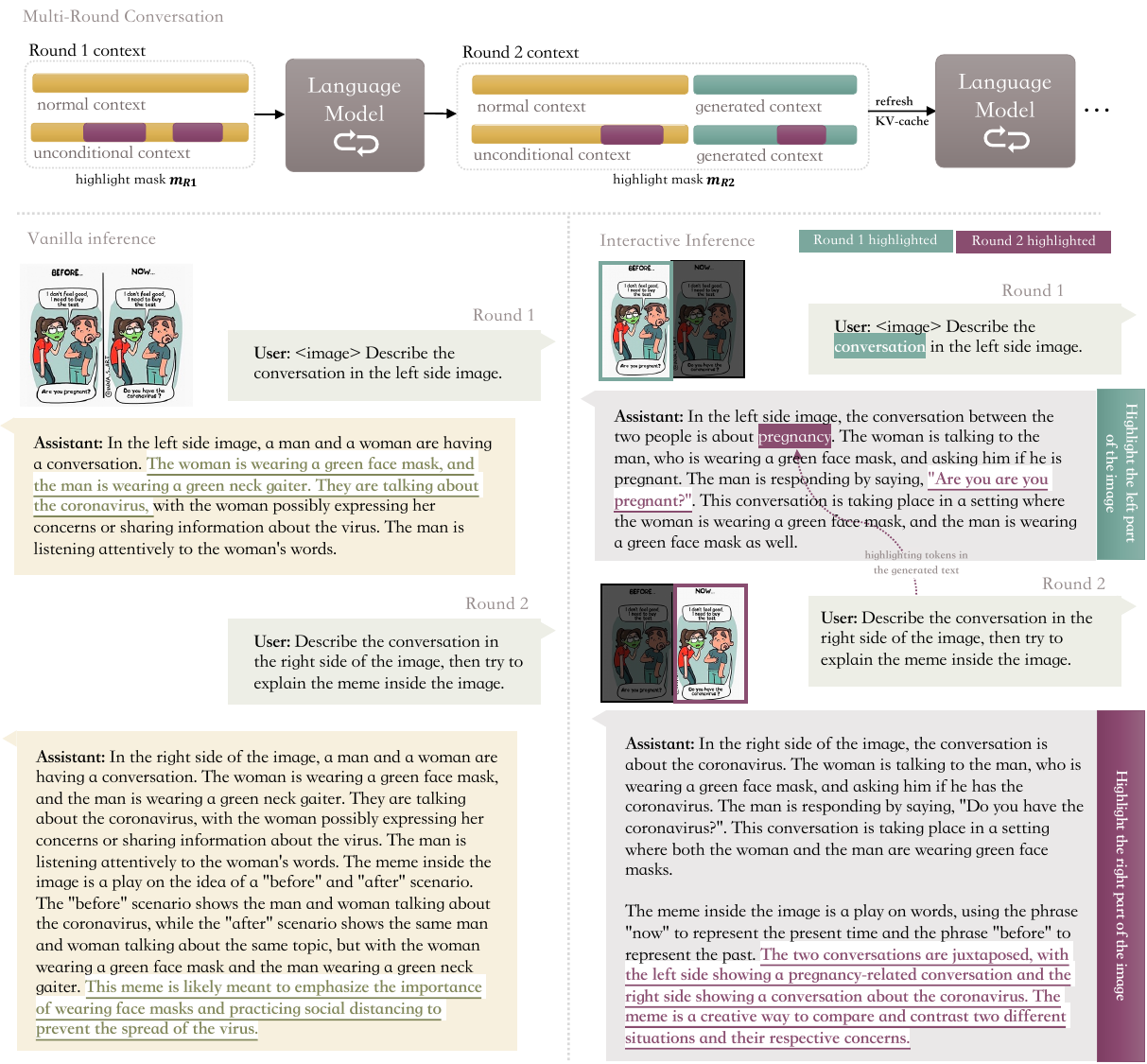}
     \caption{Multi-round interactive conversation powered by Prompt Highlighter. We illustrate the multi-round interactive conversation pipeline on the \textbf{\textit{top}}. We provide a comparison in a multi-round conversation between the vanilla inference~(\textbf{\textit{left}}) and Prompt Highlighter~(\textbf{\textit{right}}). In this example, the user highlights different contexts in two rounds.}\label{fig:supp_multi_round}
\end{figure*}

Up to this point, the examples and experimental scenarios we've showcased are for single-round dialogues or QAs. As a general-purpose interactive inference pipeline, our method can also support multi-round interactions and conversations. We provide a schematic diagram of multi-round conversation at the top of~\cref{fig:supp_multi_round}. The multi-round conversations' generation can be decomposed into single-round generations with continuously growing context, allowing Prompt Highlighter to change the focus of user input in each round by updating the input mask $\mathbf{m}$ and resetting previous KV caches.

Below the pipeline, we demonstrate how to resolve the image understanding problem, presented in~\cref{fig:supp_img_hl} case 4, through multi-round conversation and human interactions. By highlighting multi-modal tokens and guiding the model to understand the image content in a segmented manner, users can ``teach'' the model to solve problems that it was previously unable to solve.
\\


\noindent{\textcolor{purple}{\textit{\textbf{Figures 15-18 are presented on the following pages~$\downarrow$}}}}

\newpage
\clearpage
{
    \small
    \bibliographystyle{unsrt}
   \bibliography{main}
}

\end{document}